\documentclass[nohyperref]{article}

\usepackage{microtype}
\usepackage{graphicx}
\usepackage{booktabs} 
\usepackage[utf8]{inputenc} 
\usepackage[T1]{fontenc}    
\usepackage{hyperref}       
\usepackage{url}            
\usepackage{booktabs}       
\usepackage{amsfonts}       
\usepackage{nicefrac}       
\usepackage{microtype}      
\usepackage{lipsum}         
\usepackage{graphicx}
\usepackage{natbib}
\usepackage{doi}
\usepackage{amsmath}
\usepackage{multirow}
\usepackage{subcaption}
\usepackage{algorithm}
\usepackage{algorithmic}

\usepackage{hyperref}

\usepackage[accepted]{icml2022}

\usepackage{amsmath}
\usepackage{amssymb}
\usepackage{mathtools}
\usepackage{amsthm}

\usepackage[capitalize,noabbrev]{cleveref}

\theoremstyle{plain}

\theoremstyle{definition}

\theoremstyle{remark}

\usepackage[textsize=tiny]{todonotes}

\icmltitlerunning{SPDY: Accurate Pruning with Speedup Guarantees}

\begin{document}

\twocolumn[
\icmltitle{SPDY: Accurate Pruning with Speedup Guarantees}

\begin{icmlauthorlist}
\icmlauthor{Elias Frantar}{ist}
\icmlauthor{Dan Alistarh}{ist,neuralmagic}
\end{icmlauthorlist}

\icmlaffiliation{ist}{IST Austria}
\icmlaffiliation{neuralmagic}{Neural Magic}

\icmlcorrespondingauthor{Elias Frantar}{elias.frantar@ist.ac.at}
\icmlcorrespondingauthor{Dan Alistarh}{dan.alistarh@ist.ac.at}

\icmlkeywords{Machine Learning, ICML}

\vskip 0.2in
]

\printAffiliationsAndNotice{}

\begin{abstract}

The recent focus on the efficiency of deep neural networks (DNNs) has led to significant work on model compression approaches, of which \emph{weight pruning} is one of the most popular. 
At the same time, there is rapidly-growing computational support for efficiently executing the unstructured-sparse models obtained via pruning. 
Yet, most existing pruning methods minimize just the number of remaining weights, i.e. the size of the model, rather than  optimizing for inference time.
We address this gap by introducing SPDY, a new compression method which automatically determines layer-wise sparsity targets achieving a desired inference speedup on a given system, while minimizing accuracy loss. SPDY is the composition of two new techniques. 
The first is an efficient and general dynamic programming algorithm for solving constrained layer-wise compression problems, given a set of layer-wise error scores.
The second technique is a local search procedure for automatically determining such scores in an accurate and robust manner.
Experiments across popular vision and language models show that SPDY guarantees speedups while recovering higher accuracy relative to existing strategies, both for one-shot and gradual pruning scenarios, and is compatible with most existing pruning approaches. 
We also extend our approach to the recently-proposed task of pruning with very little data, where we achieve the best known accuracy recovery when pruning to the GPU-supported 2:4 sparsity pattern.

\end{abstract}

\vspace{-2em}
\section{Introduction}

Increasing the efficiency of deep neural networks (DNNs) has the potential to not just reduce the cost of compute- and energy-hungry models, but also to make them more readily available and privacy-conscious, by allowing model execution  on end-devices.
There are many approaches for compressing DNNs for increased efficiency~\cite{hoefler2021sparsity, gholami2021survey}. The one we focus on in this paper is \textit{sparsity}, i.e. setting to zero a large fraction of the values in the big parameter matrices of a neural network. Sparsification of neural networks has a long history~\cite{lecun1990optimal}, and current pruning techniques can shrink models by more than an order of magnitude, while largely preserving accuracy~\cite{hoefler2021sparsity}. 

Pruning methods are usually categorized by the granularity at which they are applied: \emph{Structured pruning} aims to drop groups of related weights, e.g. filters in a convolutional layer, which directly leads to improved inference speeds, but is usually limited in the compression amount before significant accuracy loss occurs.
\emph{Unstructured pruning} methods have traditionally been focused on reducing model size, as they drop individual weights in often random patterns, which is harder to translate into faster execution. More recent \emph{semi-structured} methods~\cite{NVIDIASparse, zhou2021learning, lagunas21block} trade off additional structure in the zeroed weights, e.g. small rectangular blocks, with higher accuracy loss relative to unstructured pruning. 

On the runtime side, increasingly advanced algorithms have been introduced to provide computational speedup also for \emph{unstructured} sparse models, whether executed on CPUs \cite{pmlr-v119-kurtz20a, deepsparse, elsen2020fast}, GPUs \cite{sgk_sc2020}, or specialized hardware~\cite{han2015learning, dave2021hardware}. 
Currently, unstructured pruning provides some of the best compression-to-accuracy trade-offs among existing approaches~\cite{hoefler2021sparsity}, and efficient sparse inference on CPUs, which is our main focus, is particularly interesting in terms of accessibility and cost. Further, several commodity CPUs, e.g. current AMD models, do not support efficient quantized arithmetic,  making sparsity their primary  means of accelerating inference. 

One key practical issue is that state-of-the-art \emph{unstructured} pruning methods, e.g.~\cite{evci2020rigging, singh2020woodfisher, schwarz2021powerpropagation, peste2021ac}, do not directly take the behavior of acceleration methods into account, while existing speedup-aware \emph{structured} pruning methods are not straightforward to adapt to the unstructured case. This means that sparsifying a model to reach a certain speed, rather than a certain size, with minimal accuracy loss, is still an extremely laborious process. 

\textbf{Contribution.} We provide a general  solution to this problem, called \emph{learned efficient Sparsity Profiles via DYnamic programming search (SPDY)}. SPDY  automatically determines how much to prune each individual network layer to achieve a desired speedup on a given inference engine and hardware combination, while minimizing accuracy loss.
The underlying optimization problem solved by SPDY is very general, as it also occurs in the context of structured pruning~\cite{he2018amc}, non-uniform layer quantization \cite{hubara2021accurate, yao2021hawq}, low-rank decomposition \cite{liebenwein2021compressing}, or even gradient compression \cite{markov2021project}. 
While we focus on unstructured pruning here, as it is an under-explored area, our approach should extend to these other settings as well. 

First, unstructured pruning with a speedup target can be viewed as  constrained optimization problem in terms of layer-wise execution timings, and so-called layer-wise ``error scores,'' and we propose an essentially exact dynamic-programming solver for this problem. This algorithm is extremely efficient: it has \emph{linear complexity} in the number of layers times the number of possible sparsity choices per layer, and can be easily scaled to very large models. 

Second, we address how to reliably determine layer-wise error scores for unstructured sparsity. We first observe that known metrics, e.g. (normalized) weight magnitudes, do not correlate consistently with superior accuracy of the resulting unstructured sparse models. 
We then introduce a new approach which \emph{learns} the layer-wise error-scores automatically, based on the network's global pruning behavior on a small set of input data. This relaxes the strictly layer-wise problem and thus makes it possible to account for global cross-layer effects, while still utilizing the advantages of the original formulation.
Specifically, SPDY determines good layer-wise error scores via local search, 
which assesses the quality of profiles determined through our DP algorithm by how well a ``reconstructed'' version of the sparse model behaves on calibration data. For sparse model reconstruction, we leverage a new variant of the AdaPrune one-shot pruning technique~\cite{hubara2021accelerated}.

\begin{figure}[t]
    \centering
    \includegraphics[width=.8\linewidth]{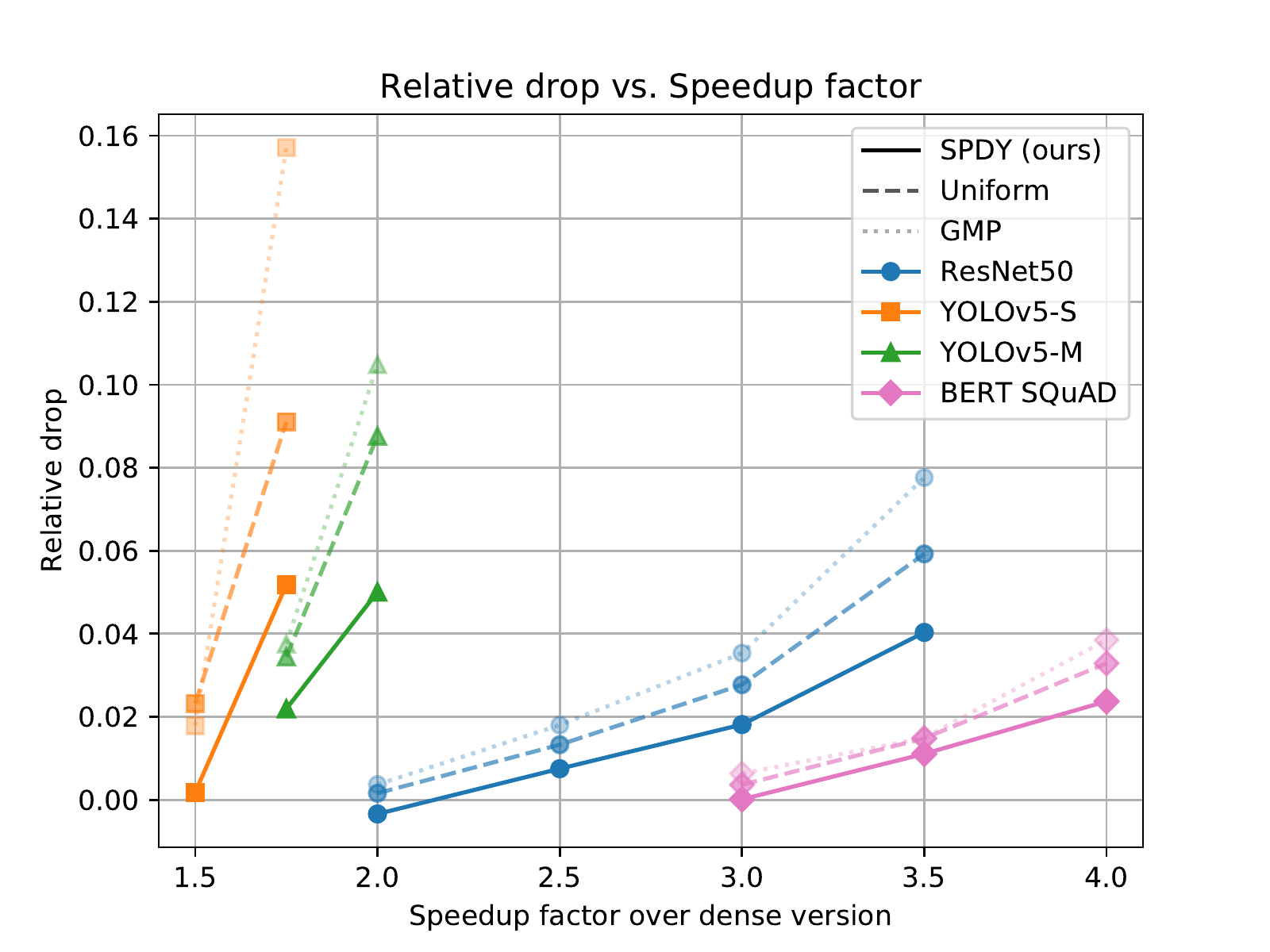}
    \vspace{-10pt}
    \caption{Speedup and relative performance measure drop trade-off after gradual pruning for SPDY and baselines on various models.}
    \label{fig:rel-drops}
   \vspace{-15pt} 
\end{figure}

We apply SPDY to determine 
``optimal'' layer-wise compression levels for a sparsity-aware CPU inference engine~\cite{deepsparse}, both for \emph{one-shot} and \emph{gradual} pruning. 
We first show that SPDY sparsity profiles can be used to compress a larger model, e.g. ResNet101, to match the inference speed of a smaller model, e.g. ResNet50, \emph{in a single compression step, without finetuning}, and still maintain an accuracy advantage.  
We then show that the layer-wise sparsity targets found by SPDY result in models with significantly better accuracy-vs-speed trade-offs than existing baselines, also when applying state-of-the-art \emph{gradual pruning} methods, as shown in Figure \ref{fig:rel-drops}. Further, SPDY can be used in conjunction with quantization-aware training. 

As a second application, we consider GPU acceleration, where we propose an enhancement of the AdaPrune method~\cite{hubara2021accelerated}, which we call \emph{global AdaPrune (gAP)}. We apply gAP to generate models with the GPU-supported 2:4 sparsity pattern in the ``post-training pruning'' setting of~\cite{hubara2021accelerated}, where only a small amount of data is available for re-calibrating the model. 
Our extension significantly outperforms the original AdaPrune: the gAP models with 2:4 sparsity have higher accuracy than AdaPrune models imposing the much less stringent 4:8 sparsity pattern.

In sum, our work introduces two new techniques for accuracy-aware acceleration of deep learning models, applicable to both CPU and GPU-based inference, which can produce state-of-the-art results in both settings. 
While we focused on unstructured and semi-structured pruning in our experiments, our approaches should be directly extensible to other settings.
We provide efficient implementations of our methods at \url{https://github.com/IST-DASLab/spdy}.

\vspace{-5pt}
\section{Related Work}

Existing pruning techniques range from simple approaches like gradual magnitude pruning \cite{hagiwara1994,zhu2017prune}, which periodically drops the fraction of the weights with lowest magnitude, followed by model finetuning, to dynamic techniques like Soft Threshold Reparametrization \cite{kusupati2020soft}, Movement Pruning \cite{2020-sanh}, or Rigging the Lottery~\cite{evci2020rigging}, which adapt mask selection during training itself. Surprisingly, properly-tuned gradual magnitude pruning is often competitive with more complex methods~\cite{singh2020woodfisher, evci2020rigging, frantar2021m, peste2021ac}. None of the above pruning techniques specifically optimize for fast inference. 
However, most can be easily modified to work with layer-wise sparsity targets, and are thus complementary to the methods we introduce; we illustrate this in our experiments by employing a range of different pruning techniques for accuracy recovery.

While structured pruning is primarily concerned with inference speed, many works focus instead on pruning FLOPs. Yet,~\citet{liu2021group} found that this is sometimes less correlated with real speedups than just reducing model size. This highlights the need to operate directly on \emph{timing data}. Related structured pruning works are Constraint-Aware Importance Estimation (CAIE) \cite{wu2020constraint} and Knapsack Pruning (KP) \cite{aflalo2020knapsack}. The former greedily ranks filters by considering their contributions towards multiple resource constraints, while the latter directly solves for the filters to select under a single constraint, a 0/1 knapsack problem, via dynamic programming. The optimal solution to this knapsack problem can be approximated with a greedy approach similar to single-constraint CAIE or \cite{yang2020automatic}. However, in our constrained optimization problem (see Section \ref{sec:optimization-problem}) the choices, i.e. the sparsity levels, are more than binary per ``item''/layer, and in practice layer speedups under sparsity are also non-linear, which makes an accurate approximation of the optimal solution more difficult. Furthermore, we show that the weight magnitudes and individual loss impact metrics, used by CAIE and KP, are not reliable for fine-grained unstructured pruning under constraints. Instead, our approach ``learns'' the error metrics automatically, allowing it to also inject global information into the layer-wise problem, which we find to be crucial for reliably finding good higher speedup solutions.

The idea of learning the layer sensitivities is related to  Automated Model Compression (AMC) \cite{he2018amc} which uses one-shot pruning performance as a proxy to train a reinforcement learning agent that predicts layer-wise (structured) sparsity targets. The idea of performing layer reconstruction via linear regression, introduced by \cite{he2017channel} and further explored by~\cite{evci2018mean}, is a precursor to the AdaPrune method, upon which our accuracy evaluation is based. 
However, we directly solve for the speedup constraint with an efficient algorithm, whereas AMC enforces the constraint only implicitly by restricting the action space of the agent. 
Yet, this and similar reinforcement learning approaches~\cite{ashok2018n2n} suffer from high tuning complexity, making them difficult to apply to new models, and have not found widespread adoption. In contrast, we demonstrate how SPDY, with a fixed set of hyper-parameters, works reliably across tasks and models.

Another related research direction is \emph{speedup-aware model quantization}, in the form of recent approaches like AdaQuant \cite{hubara2021accurate}, HAWQv3 \cite{yao2021hawq} and BRECQ \cite{li2021brecq}. These approaches solve constrained optimization problems to identify optimal per-layer bit-widths. A key simplifying feature is that these problems only have a very small number of possible choices per layer, as there are few default precision levels. Thus, the resulting problem can be solved fast without custom algorithms. This is not the case for unstructured pruning, where there are significantly more pruning choices. In addition, methods for accurate ``post training quantization'', i.e. with little available training data, like AdaQuant and AdaRound \cite{nagel2020up}, inspired the pruning equivalent AdaPrune \cite{hubara2021accelerated}. These techniques all perform a layer-wise optimization of the compressed weights to produce outputs as close as possible to the original model. 
We use AdaPrune as a basis for our one-shot pruning approach, but also extend it significantly.

In general, from the perspective of model compression under a target constraint, SPDY can be seen as a fusion of global search based approaches like AMC and layer-wise constraint-solver methods like AdaQuant; combining the respective advantages of both schemes.

\vspace{-5pt}
\section{Methods}

\subsection{Pruning for Speed: The Abstract Problem}
\label{sec:optimization-problem}

There are two core problems when trying to search for a fast \textit{unstructured} sparse model: (a) each individual weight (of which there are typically many millions) can either be dropped or kept, resulting in an enormous search space, and (b) the change in execution time for deleting a single weight is practically too small to be measured. The former challenge makes directly-extending filter-level optimization approaches \cite{aflalo2020knapsack} for structured pruning difficult, while the latter requires ranking-based approaches \cite{wu2020constraint} to rely on usually-inaccurate FLOP proxies~\cite{liu2021group} instead of actual timings.

However, both of these problems can be circumvented by leveraging the fact that the unstructured sparsity masks produced by established pruning methods have close to random structure \cite{sgk_sc2020}. Thus, unstructured acceleration techniques cannot rely on specific patterns in the layer sparsity, and have similar performance for masks of the same layer with the same sparsity level. Thus, we can reduce the overall problem of ``pruning for speed'' to identifying target sparsity values $s_\ell$ for each layer $1 \leq \ell \leq L$. Further, we can accurately estimate the runtime of such a \textit{sparsity profile} by simply imposing the corresponding sparsities with random masks, without actually running any pruning algorithms. Yet, it is critical to explicitly time different layers and sparsity levels, since acceleration rates due to unstructured sparsity are typically \emph{non-linear}: at low sparsity, sparse execution may be slower than dense execution, while speedup curves tend to flatten at high sparsities.

We wish to solve the problem of finding the ``best'' sparsity profile, in the sense of yielding the smallest possible model accuracy drop, while matching a threshold execution time $T$. To make this problem tractable, we require additional approximations. First, we assume that the overall execution time is given by the sum of the individual layer runtimes $t_{\ell}^s$ for the chosen sparsities. This is not always exact, since inference runtimes may perform layer fusion, but it is generally a good estimate, as shown in e.g.~\cite{cai2018proxylessnas}. Next, we assume that pruning a layer $\ell$ to sparsity $s$ ultimately incurs some model error $e^s_{\ell}$, and that those errors are \emph{additive}. This is a strong assumption, but one which is common in literature \cite{yao2021hawq, hubara2021accurate, aflalo2020knapsack}. Finally, we assume that the set of sparsity choices $S$ is discrete, which is reasonable in practice, as very small sparsity increments usually only lead to negligible performance gains. Under these assumptions, we can state the following constrained optimization problem:
\vspace{-5pt}
\begin{align}
    \text{min}_{s_1, \dots, s_L \in S} \, \sum_{\ell = 1}^L e^{s_\ell}_{\ell} \quad \text{s.t.}  \quad \sum_{\ell = 1}^L t_{\ell}^{s_{\ell}} \leq T.
    \label{eq:opt-problem}
\end{align}
This problem can be formulated as an integer linear program (ILP) and solved with specialized software~\cite{yao2021hawq, hubara2021accurate}. However, since each additional option per layer adds $L$ variables to the ILP, whose solving time usually increases exponentially in the number of variables, running an off-the-shelf solver would be prohibitively slow, for any model of interest, as a relatively fine-grained $S$ is needed. We note that the \textit{execution time target} $T$ corresponding to a \textit{speedup target} $X$ can be calculated as $T_\text{dense} / X - T_\text{base}$ where $T_\text{dense}$ and $T_\text{base}$ are the original dense runtime and the total runtime of operations that are unaffected by pruning, respectively.

\vspace{-5pt}
\subsection{Efficiently Solving the Optimization Problem}
\label{sec:dp-algorithm}

In its most general form, the constrained optimization problem stated in Equation (\ref{eq:opt-problem}) is NP-hard and thus (most probably) requires exponential time to solve. However, if time is integer-valued, then, as we will show, the problem is actually solvable efficiently in time $O(|S| \cdot LT)$ with $O(LT)$ memory. In our use-case, we can discretize time into $B$ buckets of width $T / B$, which means that $T = B$. Since we will be interested in large enough discretization $B$, e.g. $B = 10^4$, the inherent randomness in the individual timings will usually exceed the error incurred due to discretization, thus making this extra approximation negligible.

The key to efficiently solving the discrete version of problem~(\ref{eq:opt-problem}) is the following observation: the lowest possible error achievable in the first $\ell$ layers while taking exactly time $t$ to execute and choosing sparsity $s$ at layer $\ell$, denoted by $E_{\ell}^t(s)$, is the error caused by sparsity $s$, i.e. $e^s_{\ell}$, plus the minimum error achievable with all $\ell - 1$ previous layers while taking time exactly $t$ minus the time the choice of sparsity $s$ takes at layer $\ell$. Then, the dependence of $E_\ell^t(s)$ on $s$ can be eliminated by simply taking the minimum, leading to the following recursion:
\begin{align}
    E_\ell^t &= \text{min}_{s \in S} \, E_{\ell - 1}^{t - t^s_\ell} + e_\ell^s \\
    E_1^t &= \text{min}_{s \in S'} \, e^s_1 \,\, \text{if} \,\,  S' = \{s \, | \, t^s_1 = t \} \neq \emptyset \,\, \text{else} \,\, \infty.
\end{align}
Using dynamic programming (DP), i.e. by caching intermediate values, the final solution $\text{min}_{t \leq T} \, E^t_L$ can be computed efficiently by taking the minimum over $|S|$ options at each of the $LT$ values $E^t_\ell$. This leads to the linear memory and compute costs claimed previously. Appendix \ref{app:dp-implementation} shows a complete bottom-up DP implementation including the book-keeping to  reconstruct the optimal sparsity profile. 

In practice, this approach is highly-efficient. For example, when $T = 10^4$, $|S| = 42$ and $L = 52$ (a ResNet50 configuration) the DP algorithm finds the optimal solution in less than 100 milliseconds on a CPU and scales linearly for each parameter individually. Finally, we emphasize that this method could be used to solve any kind of layer-wise optimization problem written as (\ref{eq:opt-problem}). This includes, for instance, non-uniform layer quantization \cite{hubara2021accurate, yao2021hawq}, where the choices are quantization levels, low-rank decomposition \cite{liebenwein2021compressing}, where choices are simply ranks, or gradient compression \cite{markov2021project}, where the choices are the gradient bit-width.
The challenge for adapting the procedure to a new setting is to find a robust and accurate layer scoring metric. 
Next, we derive such a metric for unstructured pruning. 

\vspace{-5pt}
\subsection{Learning the Error Metric}
\label{sec:error-metric-learning}

For the optimal solution of optimization problem (\ref{eq:opt-problem}) to be useful, we need meaningful  error metric values $e_\ell^s$. However, especially in the context of unstructured pruning, it is quite challenging to define a general such metric, which works well across different DNNs. This is illustrated by Table~\ref{tab:metrics-problem} comparing the one-shot accuracy of profiles generated via the DP algorithm in combination with common metrics such as squared weights \cite{yao2021hawq}, squared weights normalized \cite{liu2021group} or the layer-wise loss change \cite{hubara2021accurate}. More details and additional experiments can be found in Appendix \ref{app:ablation}. While designing a robust and precise pruning error metric from first principles is an interesting direction for future work, the alternative we propose is to circumvent this issue by ``learning'' an appropriate metric automatically, which we find to consistently outperform manual options, as shown in Table \ref{tab:metrics-problem}. This also has the additional advantage that global information, e.g. if two consecutive layers should not both be heavily pruned, is integrated as well. We emphasize that this approach is only enabled by the high efficiency of our DP algorithm, since in order to reliably learn a metric, we will have to solve the constrained optimization problem a large number of times.

\begin{table}[h!]
    \centering
    \scalebox{.75}{
        \begin{tabular}{|l|c|c|c|c|c|}
            \toprule
            \multirow{2}{*}{Method} & \multicolumn{2}{c|}{ResNet34} & \multicolumn{2}{c|}{BERT} \\
            & $2.50\times$ & $3.50\times$ & $2.50\times$ & $3.50\times$ \\
            \midrule
            Uniform & 64.65 & 42.38 & 57.42 & 09.00 \\
            DP + squared & 54.18 & 31.59 & 71.20 & 07.88 \\
            DP + squared norm. & 56.32 & 09.15 & 58.77 & 06.17 \\
            DP + layer-wise loss & 65.92 & 46.11 & 30.59 & 06.13 \\
            \midrule
            \textbf{SPDY} & \textbf{68.23} & \textbf{51.68} & \textbf{74.59} & \textbf{18.83} \\
            \bottomrule
        \end{tabular}
    }
    \vspace{-5pt}
    \caption{One-shot accuracy comparison of SPDY search with DP using several common error metrics.}
    \label{tab:metrics-problem}
    \vspace{-10pt}
\end{table}

Generally, the difficulty of pruning a layer depends on how sparse a layer already is and how much of the remaining weights are to be pruned: pruning 10\% of the remaining weights will be easier than pruning 20\%, and pruning a layer that is 50\% sparse will be a lot easier than if the layer is 90\% sparse. This suggests that we should view the error in log-sparsity space, i.e. in steps of pruning $\delta$ percent of the remaining weights and, further, that the error still increases considerably faster than linearly in this parametrization. For this reason, we suggest the simple quadratic error model for each layer shown in (\ref{eq:error-approx}) where the scalar coefficient $c_\ell$ controls how quickly the error increases, i.e. it represents the sensitivity of the corresponding layer, and $i \in \{0, \dots, |S| - 1\}$. We choose a quadratic approximation because this is the ``simplest'' function with the intuitive properties discussed above. In addition, the sensitivity coefficients can be interpreted as the curvature of the errors.
\vspace{-5pt}
\begin{equation}
    \label{eq:error-approx}
    e_\ell^s = c_\ell \cdot \Big(\frac{i}{|S| - 1}\Big)^2, \quad s = 1 - (1 - \delta)^i.
\end{equation}
Given layer-wise timings $t^s_\ell$ for all sparsity levels, and using the error definition (\ref{eq:error-approx}), the DP algorithm will produce a valid profile with the desired speedup for any sensitivity coefficient vector $\mathbf{c} = (c_1, \dots, c_L) \in [0, 1]^L$. This means that we have reduced our original \emph{constrained} problem to the \emph{unconstrained} optimization problem of finding a $\mathbf{c}$ for which the DP algorithm will return a good sparsity profile with the target speedup. 

\begin{figure}[h]
    \centering
    \includegraphics[width=.7\linewidth]{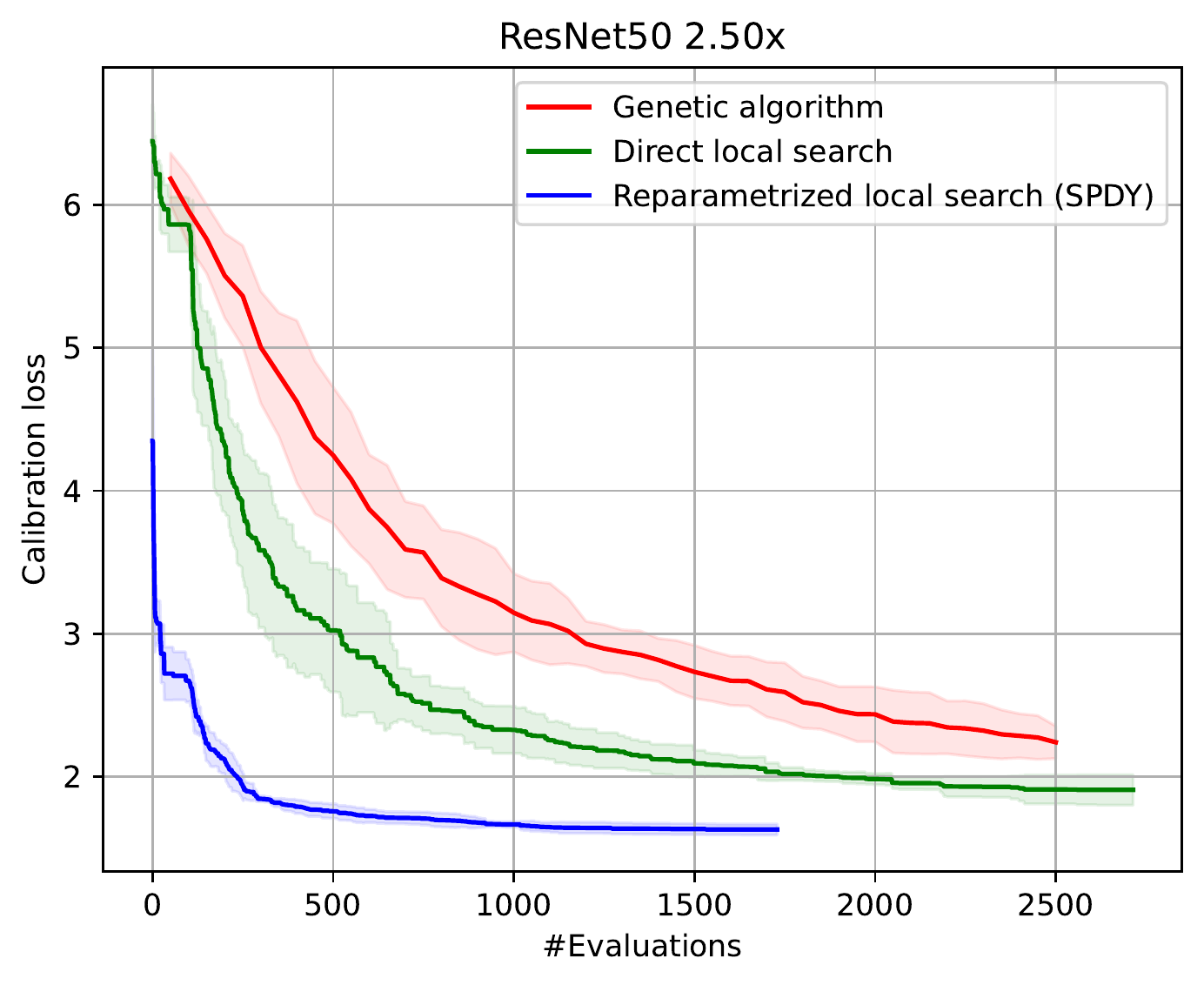}
    \vspace{-10pt}
    \caption{Comparison of SPDY with a direct search and genetic programming.}
    \label{fig:search-comparison-new}
    \vspace{-10pt}
\end{figure}

We now need an efficient procedure to determine how ``good'' a given profile is, which we discuss in Section \ref{sec:assessing-profile-quality}.  
Assuming such a procedure is available, we can then optimize $\textbf{c}$ with some heuristic optimization method, e.g. local search. It may seem that this reformulation has not lead to significant progress: the goal remains to find one choice per layer (now a sensitivity rather than a sparsity) which ultimately yields a good profile. However, we have actually eliminated the speedup constraint, which is automatically taken care of by the embedded DP algorithm. Further, various useful priors are directly enforced in a principled way, e.g. layers with poor acceleration are only pruned strongly for low sensitivity values, and  high sparsities are chosen only if they are required to reach the target speedup. As shown by comparing with a direct search (without reparametrization) and with a genetic programming method \cite{guo2020single} (Figure~\ref{fig:search-comparison-new}), the SPDY reparametrization leads to better solutions with fewer model evaluations and reduced variance. Appendix \ref{app:search} provides more details and additional analysis.

For the heuristic coefficient optimization, we have found the following \emph{randomized neighborhood-shrinking local search} to work well, and therefore use it in all our experiments.

\vspace{-5pt}
\begin{enumerate}
    \item Sample 100 random candidates for $\mathbf{c}$ and choose the one with the best resulting profile as initial $\mathbf{c}^*$.
    \vspace{-5pt}
    \item Copy $\mathbf{c}^*$ while uniformly resampling $k = \lceil 0.1 \cdot L \rceil$ random elements and replace the original $\mathbf{c}^*$ with the copy if the resulting profile is better. Stop the process if there was no improvement in 100 trials.
    \vspace{-5pt}
    \item Decrement $k$ by 1 and repeat step 2. If $k = 0$, then return the current $\mathbf{c}^*$ as the final solution.
\end{enumerate}
\vspace{-10pt}

\vspace{-5pt}
\subsection{Quickly Assessing the Quality of a Sparsity Profile}
\label{sec:assessing-profile-quality}

We are now left with the problem of efficiently evaluating the quality of a given sparsity profile, that is, predicting the accuracy of the resulting sparse model after fine-tuning.  
Intuitively, the main issue here is that, while applying a single pruning step, e.g. removing a fraction of weights by magnitude, may be very fast, pruned model accuracy collapses dramatically even at moderate sparsities, if pruning is applied in one-shot, and may not correlate well with model accuracy after fine-tuning. 
Recently, approaches such as second-order pruning~\cite{singh2020woodfisher,frantar2021m}, as well as AdaPrune (AP) \cite{hubara2021accelerated}, have dramatically improved one-shot pruning performance, mostly by  ``adapting'' the remaining unpruned weights to reduce the loss increase due to weights being removed at a step. 
We will leverage this idea to address our estimation problem, in particular via the AdaPrune approach. 

Specifically, assuming that the layer $\ell$ output function is $f_\ell(X, W)$, where $X$ are the sample inputs of a small calibration data set and $W$ are weight values, AP optimizes the sparse weights remaining after pruning $W^s$ to best reconstruct the ``functionality'' of the original dense weights $W$, independently for each layer, by minimizing the squared error between the dense and the sparse layer output:
\begin{equation}
    \label{eq:adaprune}
    \text{argmin}_{W^s} \, ||f_\ell(X, W) - f_\ell(X, W^s)||^2_2, \textnormal{ for layer $\ell$.}
\end{equation}
We leverage AdaPrune to evaluate the quality of a profile by building a \emph{layer reconstruction database}. This database stores for each layer $\ell$ and each sparsity $s$ the ``reconstructon'' of the remaining weights after sparsity $s$ has been imposed via AdaPrune. Then, we can check the quality of an arbitrary \emph{non-uniform} sparsity profile by performing two steps.
First, we query the database for the corresponding reconstructed weights of each layer, each at its target sparsity. 
Second, we ``stitch together'' the resulting model from the reconstructed weights, and evaluate it on a given small validation set. In practice, we use the same data for validation as for the AP; similar to~\cite{hubara2021accelerated}, we do not observe any overfitting.
The loss of the model on this calibration data is a proxy for the ``quality'' of the sparsity profile.

Our results show that this ``reconstruction database'' approach provides significant improvements when estimating the quality of a given profile relative to previous approaches, such as measuring the loss directly after one-shot  magnitude pruning~\cite{he2018amc}.
Yet, we found that we can still enhance its accuracy (see Appendix \ref{fig:oneshot-comparison} for experiments), especially when targeting high sparsity values, by performing this estimation \emph{iteratively}. 
Specifically, we start from the observation that solving the AP optimization problem in~(\ref{eq:adaprune}) can significantly alter the magnitudes of the unpruned weights. Practically, the weights chosen to be pruned if we applied a single high-sparsity step, e.g. 0\% to 60\%, could be quite different from those that would be selected if we pruned in several smaller increments, of e.g. 6 steps of 10\%, after each of which we update the remaining weights.
Following this intuition, our method performs iterative pruning via the AdaPrune approach, where the target sparsity is reached in several smaller pruning steps with AP optimization in between.
For the reconstruction database generation, this just means that we bootstrap the pruning process of the next sparsity level with the result of the previous one, which incurs no extra cost. Finally, we illustrate the key role of the reconstruction database for SPDY in Appendix \ref{app:ablation}.

\vspace{-5pt}
\subsection{The SPDY Method: Overview}
\label{sec:spdy-overview}

We now summarize the full SPDY method which is the combination of all the techniques discussed so far. A visual summary is given by Figure \ref{fig:SPDY-summary} and corresponding pseudo code can be found in Appendix \ref{app:pseudocode}. Our system takes as input a target execution time $T$, which is easily calculated from a target speedup $X$ (see Section \ref{sec:optimization-problem}), and a set of possible sparsity choices $S$. We then generate timing data for each layer and each sparsity choice. Additionally, we precompute a reconstruction database for fast and accurate one-shot pruning (see Section \ref{sec:assessing-profile-quality}). The output sparsity profile is then determined by a cyclic search procedure which finds sensitivity coefficients that impose a layer-wise error metric (see Section \ref{sec:error-metric-learning}). Together with the previously computed timing data, these error values are passed to a dynamic programming solver (see Section \ref{sec:dp-algorithm}) which determines a sparsity profile with target execution time $T$ and minimal total error. The actual quality of this profile is then determined by stitching together layers from the reconstruction database (see Section \ref{sec:assessing-profile-quality}) and computing the loss of the composite model on a small calibration set. The search procedure  (see Section \ref{sec:error-metric-learning}) attempts to minimize this loss by adapting the layer-wise sensitivity coefficients and ultimately outputs the sparsity profile determined by the DP algorithm when applied to the best found sensitivities.

\begin{figure*}[h]
    \centering
    \includegraphics[width=.85\textwidth]{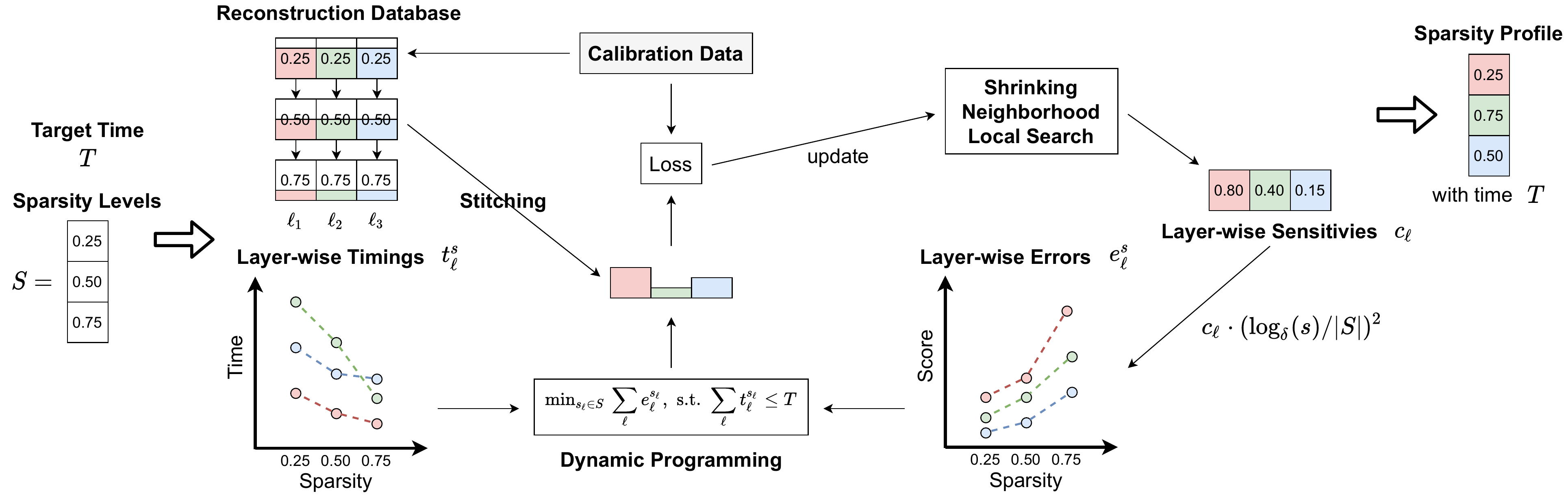}
    \vspace{-10pt}
    \caption{A visual overview of the full SPDY method.}
    \label{fig:SPDY-summary}
    \vspace{-10pt}
\end{figure*}

\vspace{-5pt}
\subsection{Post Training Pruning via global AdaPrune (gAP)}

An appealing but challenging setup is \textit{post training quantization}~\cite{nagel2020up, hubara2021accurate}, where acceleration should be done with a small amount of calibration data, without  any finetuning. 
The \emph{pruning} version of this problem is an interesting target for SPDY, since architectures such as AMD CPUs do not have quantization support, but have good speedups even at moderate sparsities.

\citet{hubara2021accelerated} showed the first \textit{post training pruning} results with acceptable accuracy drops when applied to the 4:8 semi-structured sparsity pattern using their AdaPrune (AP) approach. However, for higher sparsities, the accuracy drop of AP becomes too large to be directly useful. To address this, we introduce an extension of AP we call \textit{global AdaPrune}, or gAP for short, which boosts AP accuracy and thereby extends the practicality of post training pruning. We note that this technique is orthogonal to the SPDY method introduced in the previous sections, but can be very useful as an additional step for post-training applications.

The main motivation behind gAP is that standard AP optimizes each layer independently, thus not taking compounding errors into account, and also not considering that a slightly larger error on one layer might allow significantly reducing the error on another one. Thus, we complement AP with an optimization process over the small calibration set which (globally) optimizes the full model, in order to minimize the sum of relative layer-wise errors. We normalize the error of each layer by the squared magnitude of the output of the original dense model. This is important to ensure that all layers influence the objective equally. Specifically, let $f_\ell(X_\ell, W_\ell)$ be the output of layer $\ell$ in the original dense model and let $f_\ell(X_\ell^s, W_\ell^s)$ be the output of layer $\ell$ in the sparse model. Then, the gAP loss is written as (\ref{eq:gap-loss}) and can be minimized by gradient-based optimization: 
\begin{equation}
    \label{eq:gap-loss}
    \mathcal{L}_{\text{gAP}}(W^s) = \sum_{\ell = 1}^L \frac{||f_\ell(X_\ell, W_\ell) - f_\ell(X^s_\ell, W^s_\ell)||^2_2}{||f_\ell(X_\ell, W_\ell)||^2_2}.
\end{equation}
\vspace{-10pt}

\vspace{-5pt}
\section{Experiments}
\label{sec:experiments}

\paragraph{Setup.} We now describe our experimental setup. In all our experiments, we use the same set of sparsity targets for each layer $S = \{0\} \cup \{ 1 - (1 - 0.4) \cdot \delta^i \, | \, i = 0, \dots, 40\}$ with $\delta = ((1 - 0.99) / (1 - 0.4))^{1 / 40}$. That is, we either set a layer to dense, or to one of  41 sparsity levels, each of which prunes an additional $\approx 10\%$  of the remaining weights. 
The $0.4$  sparsity lower bound is the minimum sparsity at which the inference runtime provides some acceleration over dense execution, while $0.99$ is an upper limit to prevent model breakdown. 
For time discretization, we always use $B = 10^4$ buckets as individual units of time. 
For experiments on ImageNet~\cite{deng2009imagenet} we follow ~\cite{hubara2021accelerated}, by defining the calibration set for AP, gAP and the profile search to contain exactly one randomly-selected training image per class. For other tasks, we select 1000 training samples at random for the calibration set. The reconstruction database generation performs 10  epochs of optimization over this calibration set, using Adam \cite{kingma2014adam} with batchsize 32 and learning rate $10^{-3}$ per sparsity level while gAP runs for 100 epochs with learning rate $10^{-5}$ and frozen batch norms. The profile search follows Section \ref{sec:error-metric-learning}. With these settings, applying SPDY takes, for instance, 16min (13 database + 3 search) for ResNet18 or 51min (29 + 23) for ResNet50. This is executed on a single
NVIDIA 3090 GPU, and can be significantly optimized.

We measure speedups and execute inference on the publicly-available DeepSparse v0.9.1 CPU inference engine \cite{deepsparse, pmlr-v119-kurtz20a}, which is competitive when executing dense models with the standard ONNX and OpenVINO runtimes, but can additionally leverage unstructured sparsity for speedup. 
DeepSparse is currently the most mature CPU engine with sparsity support; yet, we emphasize that alternatives exist~\cite{elsen2020fast, sgk_sc2020}, and that our techniques are independent of the runtime.

We compare against two baseline profiles: the \emph{uniform} profile of minimal sparsity which matches the target execution time, and the profile determined by pruning using \emph{global magnitude pruning (GMP)} until the target speedup is reached. 
Both are direct speedup-aware adaptations of widely-used pruning strategies, which result in reasonable profiles without manual tuning \cite{he2019filter, liu2021group}.
Additionally, following other works \cite{elsen2020fast, jayakumar2021top}, we always keep the first and the last layer dense, since those typically have disproportionately large effects on the model accuracy while taking only a very small fraction of the total computation time. (The only exception is YOLO, where the 3 output layers are compute-heavy, so we only skip the input.) We emphasize that this helps the baseline profiles (uniform and GMP) significantly. (For instance, \citet{peste2021ac} report 73.14\% accuracy for a 95\% sparse ResNet50  versus 74.16\% for one with first and last layer skipped.) This choice has little effect on SPDY since our method generally detects the compute/sensitivity imbalance automatically, and skips those layers. We also experimented with a GMP version that normalizes magnitude scores by the corresponding FLOPs. However, this did not provide improvements without model-specific manual tweaking of layer-wise normalizers, e.g. on ResNet50 it would prune all the early layers to 99\% sparsity. To obtain a consistent AP reconstruction database and timing data, we round our baselines to the fixed sparsities in $S$.

\begin{figure}[h]
    \vspace{-5pt}
    \centering
    \includegraphics[width=.8\linewidth]{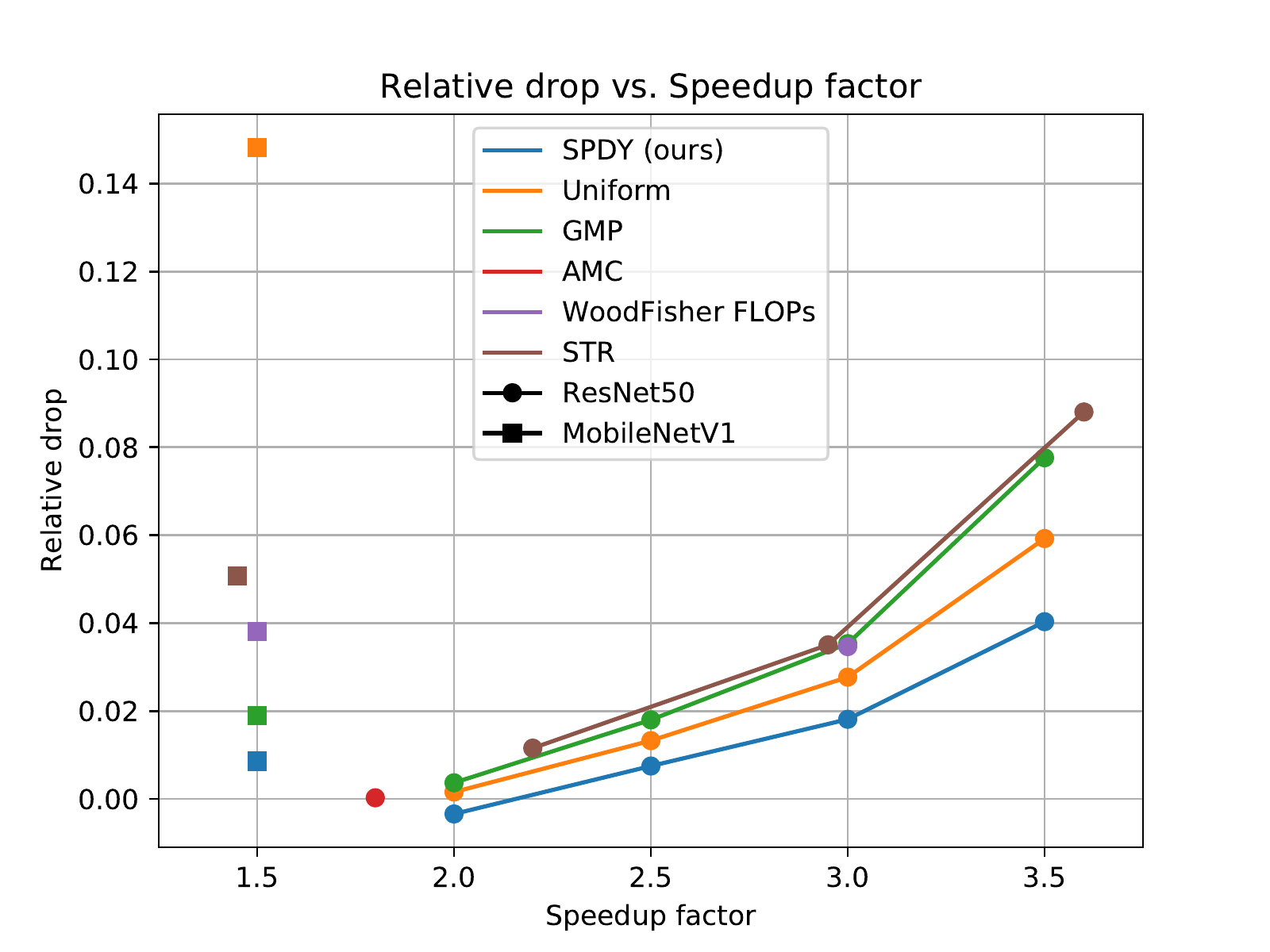}
    \vspace{-10pt}
    \caption{Comparison with results of speedup-related unstructured pruning methods on ResNet50 and MobileNetV1 in terms of relative accuracy drop.}
    \label{fig:other-comp}
    \vspace{-15pt}
\end{figure}

There are some unstructured pruning methods which relate their results to speed/efficiency: STR \cite{kusupati2020soft}, WoodFisher FLOPs \cite{singh2020woodfisher} and AMC \cite{he2018amc}. The comparison in Figure \ref{fig:other-comp} shows that they all perform similar or worse than our simpler uniform and GMP baselines. Additionally, some of these methods do not provide  mechanisms to control the target speedup. Thus, we focus on comparisons with uniform and GMP profiles.

Layer-wise timings for the AMD system are collected on an Amazon AWS c5a.8xlarge machine with 16 cores, while for Intel CPUs we use a c5.9xlarge server with 18 cores. 
The listed speedups are for batchsize 64, except for BERT \cite{devlin2018bert}, which uses batchsize 16. 
The speedups given in the tables below are calculated in terms of the layer-wise timing data, which usually slightly underestimates the real speedups when running the engine with all extra optimizations turned on. This allows comparing different profile generation methods in isolation of highly engine-specific timing inaccuracies and will also make it easy for future work to directly compare with our results using the published timing data. We provide real speedup information for our most interesting profiles in Appendix \ref{app:real-timings}, which demonstrates that the layer-wise approximation is indeed quite accurate in most cases.

\textbf{Post Training Pruning.}
We begin in the post-training pruning setting, and consider the torchvision \cite{marcel2010torchvision} implementation of ResNets~\cite{he2016deep} running on an AMD CPU. These models are popular, and they can be well-accelerated by DeepSparse. 
We report performance directly after stitching the model from the AP database (labelled AP) as well as after a subsequent run of global AP (labelled +gAP). We consider the use-case of pruning each network to approximately the same inference speed as the next smallest variant available in torchvision (e.g. ResNet101 to ResNet50, see Table \ref{tab:oneshot-next}).

\begin{figure*}
    \begin{minipage}[c]{0.55\textwidth}
        \centering
        \scalebox{.75}{
            \begin{tabular}{|l|l|c|c|c|c|c|c|c|c|}
                \toprule
                \multirow{2}{*}{Base} & \multirow{2}{*}{Pruning} & \multicolumn{2}{c|}{SPDY} & \multicolumn{2}{c|}{Uniform} & \multicolumn{2}{c|}{GMP} \\
                & & AP & +gAP & AP & +gAP & AP & +gAP \\
                \midrule
                69.76 RN18 & $1.80 \times$ RN34 & \textbf{69.84} & \textbf{71.34} & 68.09 & 71.10 & 63.78 & 66.83 \\
                73.31 RN34 & $1.60 \times$ RN50 & \textbf{74.39} & \textbf{75.01} & 71.60 & 74.05 & 72.41 & 73.88 \\
                76.13 RN50 & $1.80 \times$ RN101 & \textbf{76.00} & \textbf{76.61} & 73.55 & 75.64 & 75.42 & 76.32 \\
                77.37 RN101 & $1.45 \times$ RN152 & \textbf{77.49} & 77.82 & 76.46 & 77.40 & 77.42 & \textbf{77.91} \\
                76.13 RN50 & $2.45 \times$ RN50w2 & \textbf{75.32} & \textbf{77.08} & 67.47 & 74.51 & 69.78 & 73.64 \\
                \bottomrule
            \end{tabular}
        }
        \vspace{-5pt}
        \captionof{table}{Comparing profiles when post training pruning ResNet models to the same speed as their next smallest dense counter-part.}
        \label{tab:oneshot-next}
    \end{minipage}
    \hfill
    \begin{minipage}[c]{0.40\textwidth}
        \centering
        \scalebox{.75}{
            \begin{tabular}{|l|c|c|c|c|c|c|c|}
                \toprule
                \multirow{2}{*}{Model} & \multirow{2}{*}{Dense} & \multicolumn{2}{c|}{2:4} & \multicolumn{3}{c|}{4:8} \\
                & & AP & \textbf{+gAP} & AP & A+B & \textbf{+gAP} \\
                \midrule
                RN18 & 69.76 & 67.64 & \textbf{68.76} & 68.32 & 68.63 & \textbf{69.05} \\
                RN34 & 73.31 & 71.81 & \textbf{72.59} & 72.33 & 72.36 & \textbf{72.84} \\
                RN50 & 76.13 & 73.61 & \textbf{75.00} & 74.44 & 74.75 & \textbf{75.43} \\
                RN101 & 77.37 & 75.70 & \textbf{76.75} & 76.23 & 76.48 & \textbf{76.88} \\
                RN152 & 78.31 & 76.75 & \textbf{77.75} & 77.36 & -- & \textbf{77.91} \\
                \bottomrule
            \end{tabular}
        }
        \vspace{-5pt}
        \captionof{table}{AdaPrune (AP) and global AdaPrune (gaP) for post training pruning to 2:4 and 4:8 sparsity.}
        \label{tab:rn-nm}
    \end{minipage}
    \vspace{-5pt}
\end{figure*}

Table \ref{tab:oneshot-next} clearly shows that ResNets post training pruned with SPDY and global AdaPrune always exceed the accuracy of the smaller dense variant with approximately the same speed. In several cases, the difference is  significant, e.g. $\approx 1.7\%$ Top-1 for RN34 or $\approx 1.5\%$ for RN18. Further, the SPDY profile almost always performs best. The only exception is the very large ResNet-152 with a low speedup target, where GMP performs $\approx 0.1\%$ better. Meanwhile, the SPDY profile outperforms the baselines by $\approx 1\%$ on RN50 and even by $\approx 2.5\%$ on RN50 with doubled width (RN50w2). We get similar results when pruning all models to the same $1.5\times$ speedup (see Appendix \ref{app:additional-experiments}) but with smaller relative accuracy drops. Note that uniform pruning  outperforms GMP at smaller models and higher speedups, while the situation is reversed for larger models and lower speedups.

\textbf{Compounding Pruning and Quantization.}
Quantization is an alternative compression technique, which is popular with both GPUs and end devices~\cite{gholami2021survey}, as well as high-end Intel CPUs. The DeepSparse runtime 
supports both techniques with speedup in complementary fashion. To leverage this, we set a pruning speedup target of 2x for ResNet50, and then run SPDY vs. uniform pruning in a post-training setting, followed by  Quantization Aware Training (QAT) \cite{nagel2021white} for 6 epochs to quantize  weights and recover additional accuracy. 
Quantization provides an additional $2.6\times$ speedup, so the end model is $5.2\times$ faster than the dense full-precision version, on an AWS c5.12xlarge instance. SPDY's accuracy improvement after pruning is clear: we have 73.33\% vs. 71.94\% top-1 accuracy, for the semi-structured pruning pattern with blocks of 4 consecutive zeros required by DeepSparse. This directly leads to improved 75.00\% vs. 74.33\% accuracy after the quantization-aware finetuning stage, in favor of SPDY.

\textbf{Post Training N:M Sparsity.}
To gain additional insight into the accuracy difference between our global layer reconstruction approach relative to existing methods, we investigate the performance of global AdaPrune in the context of post training pruning models to full N:M sparsity, a type of semi-structured sparsity that can be well accelerated on the newer NVIDIA GPUs. \citet{hubara2021accelerated} were the first to demonstrate that achieving N:M sparsity in the post training setting with acceptable accuracy loss is possible using AdaPrune and batchnorm tuning (A + B). Unfortunately, their results are somewhat theoretical, as they only targeted 4:8 sparsity, which to our knowledge is not yet supported by hardware. We now show that by applying global AP (after standard AP), we can exceed their 4:8 accuracy results with the significantly more challenging 2:4 sparsity pattern that is already well supported by current NVIDIA GPUs. A comparison for the ResNet model family can be found in Table \ref{tab:rn-nm}. We note that further improvements might be possible through optimized channel permutations \cite{pool2021channel}. Additionally, in Appendix \ref{app:additional-experiments}, we present results for post training 2:4 pruning YOLOv5 models and BERT.

\textbf{Gradual Pruning Results.}
\label{sec:gradual-pruning}
We now present our main results, generating speedup profiles  and then performing \emph{gradual pruning} using state-of-the-art pruning methods under their default parameters, with these layer-wise sparsity targets. We consider the standard ResNet50 model \cite{he2016deep}, as well as the hard-to-compress MobileNetV1 architecture~\cite{howard2017mobilenets, kusupati2020soft}, the Small and the Medium versions of the popular YOLOv5\footnote{We evaluate using the author's validation script with default parameters and report the pycocotools mAP@0.5.} \cite{yolov5} object detector, and the widely used  BERT-base for language modelling \cite{devlin2018bert} on the SQuAD dataset \cite{rajpurkar2016squad}. ResNet50 is gradual pruned with AC/DC \cite{peste2021ac} for 100 epochs, MobileNetV1 with M-FAC \cite{frantar2021m} for 100 epochs and YOLOv5 and BERT with gradual magnitude pruning (using the setup of \cite{kurtic2022optimal} but performing a pruning step every $0.01$ epochs) for 240 and 30 epochs, respectively. The results are summarized in Table \ref{tab:gradual-results}.

\begin{table}[h]
    \begin{minipage}[c]{\linewidth}
        \centering
        \scalebox{.75}{
            \begin{tabular}{|l|c|c|c|c|c|c|}
                \toprule
                Model & Dense & Speed. & CPU & \textbf{SPDY} & Uni. & GMP \\
                \midrule
                ResNet50 & 76.13 & $2.00\times$ & AMD & \textbf{76.39} & 76.01 & 75.85 \\
                ResNet50 & 76.13 & $2.50\times$ & AMD & \textbf{75.56} & 75.12 & 74.76 \\
                ResNet50 & 76.13 & $3.00\times$ & AMD & \textbf{74.75} & 74.02 & 73.44 \\
                ResNet50 & 76.13 & $3.50\times$ & AMD & \textbf{73.06} & 71.62 & 70.22 \\
                \midrule
                MobileNetV1 & 71.95 & $1.50\times$ & Intel & \textbf{71.38} & 61.33 & 70.63 \\
                \midrule
                YOLOv5s & 56.40 & $1.50\times$ & Intel & \textbf{55.90} & 54.70 & 55.00 \\
                YOLOv5s & 56.40 & $1.75\times$ & Intel & \textbf{53.10} & 50.90 & 47.20 \\
                YOLOv5m & 64.20 & $1.75\times$ & Intel & \textbf{62.50} & 61.70 & 61.50 \\
                YOLOv5m & 64.20 & $2.00\times$ & Intel & \textbf{60.70} & 58.30 & 57.20 \\
                \midrule
                BERT SQuAD & 88.54 & $3.00\times$ & Intel & \textbf{88.53} & 88.22 & 87.98 \\
                BERT SQuAD & 88.54 & $3.50\times$ & Intel & \textbf{87.56} & 87.23 & 87.22 \\
                BERT SQuAD & 88.54 & $4.00\times$ & Intel & \textbf{86.44} & 85.63 & 85.13 \\
                BERT SQuAD$^*$ & 88.54 & $4.00\times$ & Intel & \textbf{87.14} & 86.37 & 86.39 \\
                \bottomrule
            \end{tabular}
        }
        \vspace{-5pt}
        \captionof{table}{Comparing accuracy metrics for  sparsity profiles after gradual pruning models with respective state-of-the-art methods.}
        \label{tab:gradual-results}
    \end{minipage}
    \vspace{-20pt}
\end{table}

The SPDY profiles appear to yield more accurate final weights across all models and speedups, often with a large gap to the next best method. At higher speedups, the advantage of SPDY seems to increase further, with $1.4$ points better accuracy for ResNet50 and even $> 2$ points better mAP@0.5 for YOLO models. In the case of YOLO, it should also be mentioned that the profile generation optimizes not the relatively complex training loss but simply the squared error of the final output to the one of the original dense model. This suggests that our method can be applied successfully even if the  calibration set is \emph{unlabelled}. 

In some experiments, the improvement via SPDY is relatively small, e.g. for the lower speedup BERT runs we get only $\approx 0.3$ higher F1. However, we note that a) the performance improvement of SPDY is larger than the difference between uniform and GMP in those cases and b) this occurs after extensive finetuning, which usually narrows accuracy gaps. 
The fact that these differences are always consistent can be seen as an indication in favor of our method. 

Finally, in these experiments all profiles were generated initially, based on the dense model. Potentially, results could be improved further by rerunning SPDY from a model that has already been pruned to an intermediate sparsity. 
We test this by generating new $4\times$ speedup profiles based on the gradually-pruned $3\times$ BERT models which are also the starting points for subsequent gradual pruning (indicated by ``BERT SQuAD$^*$'' in the table). Although the absolute gap seems to stay the same, all models are noticeably more accurate, so the relative gap in terms of F1-drop is bigger, suggesting that iterating SPDY can bring additional gains.

We also highlight the ResNet50 results, achieved with AC/DC, a method that trains and prunes the model \textit{from scratch}. This implies that the sparsity profiles found by SPDY are not just fitted to a particular set of dense weights but also possess some architecture-specific generalization properties. In that context, the results of our experiments reinforce the   idea that better one-shot performance is a good indicator of performance after finetuning~\cite{he2018amc}. They seem to suggest that good sparsity profiles can ``transfer'' also when running gradual pruning, which may be loosely connected to ``lottery ticket'' approaches~\cite{frankle2018lottery}. We discuss some of the profiles produced by SPDY in more detail in Appendix \ref{app:profiles}. In general, SPDY profiles achieve the same speedup with a slightly lower overall sparsity than uniform and GMP, due to prioritizing execution time (see also Appendix \ref{app:sparsity}), which likely contributes to the higher accuracies after extensive finetuning. However,  our approach can be adapted to any additive constraint, such as parameters, FLOPs, or energy consumption.

\vspace{-5pt}
\section{Conclusion \& Future Work}
\vspace{-3pt}

We introduced SPDY, a method for automatically determining sparsity profiles optimized for a particular acceleration setup. The resulting profiles consistently perform better than ones determined by conventional methods,  without any model-specific parameter tuning,  both in one-shot and gradual-pruning scenarios. An extension of our layer reconstruction method can also provide improvements for  post-training 2:4 pruning via global AdaPrune. 

Our ideas can be extended to other settings: (1) a combination of a dynamic program constraint solver and automated learning of sensitivity scores could also be applied to determine compression profiles for low-rank approximation; (2) it could be interesting to apply our techniques to neural architecture search super-networks as a replacement for our one-shot pruning reconstruction database. In sum, our techniques provide notable improvements relative to current methods, and have significant potential for follow-up work.

\vspace{-5pt}
\section{Acknowledgements}
\vspace{-3pt}

We gratefully acknowledge funding from the European Research Council (ERC) under the European Union’s Horizon 2020 programme (grant agreement No 805223 ScaleML), as well as computational support from AWS EC2. We  thank Eldar Kurtic for code and hyper-parameters for BERT pruning, and the Neural Magic Team, notably Michael Goin and Mark Kurtz, for support with their software.

\bibliographystyle{icml2022}

\clearpage

\section{DP Algorithm Implementation}
\label{app:dp-implementation}

\begin{algorithm}[h!]
    \caption{We efficiently compute the optimal layer-wise sparsity profile with execution time at most $T$ given $S$, $e^s_\ell$, $t^s_\ell$ and assuming that time is discretized, using bottom-up dynamic programming.}
    \label{alg:dp}
    \begin{algorithmic}
        \STATE $\mathbf{D} \gets L \times (T + 1)$ matrix filled with $\infty$
        \STATE $\mathbf{P} \gets L \times (T + 1)$ matrix
        \FOR {$s \in S$}
            \IF {$e^s_1 < \mathbf{D}[1,t^s_1]$}
                \STATE $\mathbf{D}[1,t^s_1] \gets e^s_1$; \, $\mathbf{P}[1,t^s_1] \gets s$
            \ENDIF
        \ENDFOR
        \FOR {$\ell = 2, \dots, L$}
            \FOR {$s \in S$}
                \FOR {$t = t^s_\ell + 1, \dots, T$}
                    \IF {$e^s_\ell + \mathbf{D}[\ell - 1, t - t^s_\ell] < \mathbf{D}[\ell, t]$}
                        \STATE $\mathbf{D}[\ell, t] \gets e^s_\ell + \mathbf{D}[\ell - 1, t - t^s_\ell]; \, \mathbf{P}[\ell, t] \gets s$
                    \ENDIF
                \ENDFOR
            \ENDFOR
        \ENDFOR
        \STATE {$t \gets \text{argmin}_t \, \mathbf{D}[L, t]$} // return $\mathbf{D}[L, t]$ as optimal error 
        \FOR {$\ell = L, \dots, 1$}
            \STATE {$s \gets \mathbf{P}[\ell, t]$ // return $s$ as optimal sparsity for layer $\ell$}
            \STATE {$t \gets t - t^s_\ell$}
        \ENDFOR
    \end{algorithmic}
\end{algorithm}

We note that, in practice, the innermost loop over $t$ can easily be vectorized (with corresponding acceleration through specialized CPU / GPU code) via shifting of $\mathbf{D}[\ell - 1, :]$ by $t^s_\ell$, thereby making the overall algorithm highly efficient even for very fine discretization.

\section{SPDY Pseudocode}
\label{app:pseudocode}

This section provides additional details about the overall SPDY framework described in Section \ref{sec:spdy-overview} in form of pseudo code. Specifically, Algorithm \ref{alg:collect-timings} and Algorithm \ref{alg:create-db} illustrate the initial timing collection and iterative AdaPrune reconstruction database generation. Meanwhile, Algorithm \ref{alg:spdy-search} presents the SPDY search for the optimal coefficients $\mathbf{c^*}$.

\begin{algorithm}[h!]
    \caption{Collect layer-wise timings $t^s_\ell$.}
    \label{alg:collect-timings}
    \begin{algorithmic}
        \FOR {$\ell = 1, \dots, L$}
            \FOR {$s \in S$}
                \STATE $W_\ell^s \gets$ randomly prune $W_\ell$ to sparsity $s$
                \STATE $t_\ell^s \gets$ collect timing for layer $\ell$ using weights $W_\ell^s$
            \ENDFOR
        \ENDFOR
    \end{algorithmic}
\end{algorithm}
\begin{algorithm}[h!]
    \caption{Generate reconstruction database entries $W^s_\ell$.}
    \label{alg:create-db}
    \begin{algorithmic}
        \STATE $\mathbf{s} \gets$ sorted vector of sparsities in $S$
        \FOR {$\ell = 1, \dots, L$}
            \STATE $W^{s_0}_\ell \gets$ dense weights $W_L$
            \FOR {$i = 1, \dots |S|$}
                \STATE $W_\ell^{s_i} \gets$ AdaPrune for target $s_i$ with weights $W_\ell^{s_{i - 1}}$
            \ENDFOR
        \ENDFOR
    \end{algorithmic}
\end{algorithm}
\begin{algorithm}[h!]
    \caption{SPDY search for optimal sensitivity values $\mathbf{c^*}$. We use $k = 100$ and $\delta = 0.1$ in our experiments.}
    \label{alg:spdy-search}
    \begin{algorithmic}
        \STATE \textbf{function} $eval(\mathbf{c})$
        \STATE \quad $e^s_\ell \gets$ compute by formula (\ref{eq:error-approx}) using $\mathbf{c}$ for all $\ell$
        \STATE \quad $s_\ell \gets$ run DP algorithm with $e^s_\ell$ for all $\ell$
        \STATE \quad $M \gets $ stitch model for $s_\ell$ from database
        \STATE \quad Return calibration loss of $M$.
        \vspace{5pt}
        \STATE $\mathbf{c^*} \gets$ sample uniform vector in $[0, 1]^L$
        \FOR {$k$ times}
            \STATE $\mathbf{c} \gets$ sample uniform vector in $[0, 1]^L$
            \IF {$eval(\mathbf{c}) < eval(\mathbf{c^*})$}
                \STATE $\mathbf{c^*} \gets \mathbf{c}$
            \ENDIF
        \ENDFOR
        \FOR {$d = \lceil \delta \cdot L \rceil, \dots, 1$}
            \FOR {$k$ times}
                \STATE $\mathbf{c} \gets \mathbf{c^*}$
                \STATE Randomly resample $d$ items of $\mathbf{c}$ in $[0, 1]$
                \IF {$eval(\mathbf{c}) < eval(\mathbf{c^*})$}
                    \STATE $\mathbf{c^*} \gets \mathbf{c}$
                \ENDIF
            \ENDFOR
        \ENDFOR
    \end{algorithmic}
\end{algorithm}

\section{One-Shot Comparison}
\label{app:one-shot-comparison}

\begin{figure}[h!]
    \centering
    \includegraphics[width=.9\linewidth, trim={0 0 0 25},clip]{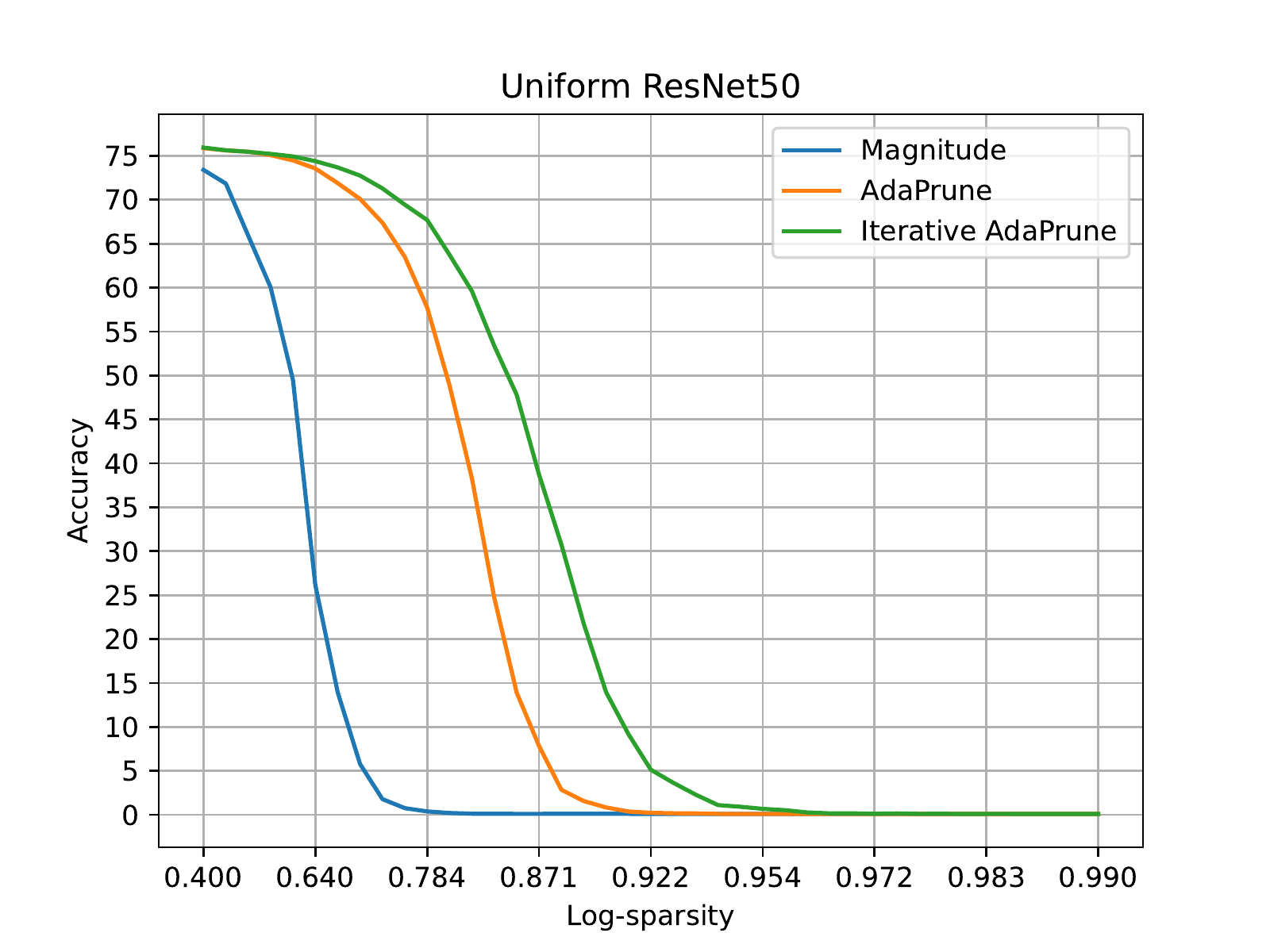}
    \vspace{-10pt}
    \caption{Comparison of magnitude pruning, AdaPrune and iterative AdaPrune in terms of one-shot performance.}
    \label{fig:oneshot-comparison}
\end{figure}

Figure \ref{fig:oneshot-comparison} demonstrates how iterative AdaPrune significantly improves over standard AdaPrune and GMP.

\begin{table*}[h]
    \begin{minipage}[c]{\linewidth}
        \centering
        \scalebox{.75}{
            \begin{tabular}{|l|c|c|c|c|c|c|c|}
                \toprule
                & \multicolumn{2}{c|}{ResNet34} & \multicolumn{2}{c|}{ResNet50} & \multicolumn{2}{c|}{BERT SQuaD} \\
                & $2.00\times$  & $3.00\times$ & $2.00\times$ & $3.00\times$ & $2.50\times$ & $3.50\times$ \\
                \midrule
                DP + squared weights & 54.18 & 31.59 & 47.44 & 03.82 & 71.20 & 07.88 \\
                DP + squared weights normalized & 56.32 & 09.15 & 35.42 & 01.20 & 58.77 & 06.17 \\
                DP + custom norm based & 66.64 & 49.17 & 66.82 & 11.11 & 72.34 & 13.64 \\
                DP + layer-wise loss / no reconstruction & 54.51 & 10.32 & 66.49 & 08.85 & 53.98 & 06.85\\
                DP + layer-wise loss / iterative AP & 65.92 & 46.11 & 69.26 & 20.22 & 30.59 & 06.13 \\
                SPDY / no reconstruction & 53.38 & 22.18 & 64.59 & 00.67 & 62.41 & 07.58 \\
                \textbf{SPDY} & \textbf{68.23} & \textbf{51.68} & \textbf{71.17} & \textbf{26.81} & \textbf{74.59} & \textbf{18.83} \\
                \midrule
                Uniform & 64.65 & 42.38 & 64.65 & 05.84 & 57.42 & 09.00 \\
                Global magnitude & 39.41 & 22.80 & 62.89 & 01.76 & 58.06 & 06.85 \\
                \bottomrule
            \end{tabular}
        }
        \vspace{-5pt}
        \captionof{table}{A comparison of one-shot accuracies for profiles generated by DP in combination with various metrics, SPDY and uniform / global magnitude pruning.}
        \label{tab:metrics-ablation}
    \end{minipage}
\end{table*}

\section{Ablation Studies}
\label{app:ablation}

In Section \ref{sec:error-metric-learning}, we briefly discussed the difficulty of designing a reliable error metric for the use in conjunction with our DP algorithm. We now present a more detailed investigation of this topic including additional experimental results. Concretely, we consider the following 5 error metrics:

\begin{itemize}

    \item \textbf{Squared Weights} --- Perhaps the most obvious error score candidate is a natural extension of the popular magnitude pruning criterion \cite{zhu2017prune}, the sum of the squared magnitudes of all pruned weights. It can also be interpreted as the squared norm of the weight perturbation, which has been used in the context of quantization \cite{yao2021hawq}.
    
    \item \textbf{Squared Weights Normalized} --- A potential problem of the previous metric is that smaller layers, which are often quite sensitive, might be pruned too strongly as their sums are inherently smaller than those of bigger layers. A simple way to address this is to normalize each sum by the number of elements in the corresponding layer. Such normalization is for example used for structured pruning in \cite{liu2021group}.
    
    \item \textbf{Custom Norm Based} --- Another possible shortcoming of both metrics mentioned so far is that the sums for high sparsity choices generally differ only in a small number elements and can thus be quite similar. This means the fact that pruning typically becomes a lot more difficult at high sparsities may not be very well reflected in those error metrics. Hence, we developed a custom norm based metric to address this. Let $\text{max}_{w \not \in W^s} \, |w|$ be the magnitude of the largest dropped weight when pruning a layer to sparsity $s$, then the error is defined as:
    
    \begin{equation*}
        e^s = \frac{\text{max}_{w \not \in W^s} \, |w|}{1 - s}.
    \end{equation*}
    
    The numerator ensures that the influence of the weight magnitudes does not decrease for high sparsities and the division by the remaining density that the error always increases strongly as a layer gets very sparse.
    
    \item \textbf{Layer-wise Loss No Reconstruction} --- One more obvious option for a layer-wise error is simply the change in loss on a small calibration set that is the result of pruning exactly this layer to a certain sparsity. This metric measures the loss after standard magnitude pruning, without any layer-wise reconstruction, similar to \cite{li2016pruning}.
    
    \item \textbf{Layer-wise Loss Iterative AP} --- Finally, we combine the layer-wise loss with the techniques we introduce in Section \ref{sec:assessing-profile-quality}, i.e. using iterative AdaPrune to reconstruct the pruned layer before measuring the loss. This is similar in spirit to the AdaQuant loss used as an error metric for quantization \cite{hubara2021accurate}.
    
\end{itemize}

We now generate profiles using DP in combination with each of the 5 described metrics, for 3 different models at 2 speedups each, a lower one where AP pruning still gives reasonable accuracies and a higher one which could be used as the starting point for extended gradual pruning as in Section \ref{sec:gradual-pruning}. We consider two ResNets to study the consistency of metrics for models of the same family and a very different BERT model. For computational reasons, we consider the accuracy after one-shot pruning by stitching together layers from the reconstruction database described in Section \ref{sec:assessing-profile-quality}, which \cite{he2018amc} as well as our main experimental results show to be decent proxies for post finetuning / gradual pruning accuracy. We compare with SPDY as well as the standard uniform and global magnitude strategies (see Section \ref{sec:experiments}), input and output layers are always skipped. All results are summarized in Table \ref{tab:metrics-ablation}.

First, we can see that the squared weights based metrics fail to beat the best non-DP profile in all but one case. The same is true for the layer-wise loss without any reconstruction. Clearly, those metrics are unsuitable for generating effective sparsity profiles. Meanwhile, the layer-wise loss with iterative AP reconstruction seems to perform quite well relative to the non-DP results on the ResNet models but performs poorly on BERT where the losses are apparently so misleading that it is beaten even by the no-reconstruction version. A closer look reveals that the layer-wise losses do not all increase strongly with higher sparsities on this large model, which leads to very aggressive pruning of several layers compounding to large accuracy drops (that are not properly reflected in the single-layer losses). Additionally, it should be noted that the layer-wise loss works better than the custom metric on ResNet50 but worse on ResNet34. This means the best performing metric does not even seem to be consistent within a single model family. In general, the custom metric appears to perform reasonably in all 6 scenarios, outperforming the non-DP profiles, sometimes even by sizable margins. However, there is still a significant gap to the SPDY results in several cases. In fact, we found that the custom metric profiles often only lead to minimal 0.1 -- 0.2 improvements over uniform ones when running extensive state-of-the-art gradual pruning, meaning that the extra improvements of SPDY are crucial to achieve the notable gaps we do in Section \ref{sec:gradual-pruning}. Overall, these experiments demonstrate that manually designing a reliable layer-wise error metric for unstructured pruning is indeed a quite challenging problem, as the automatically learned error metric of SPDY consistently beats hand-crafted metrics of varying complexity.

Finally, we also briefly study the importance of the enhanced one-shot pruning performance through our reconstruction database. For that purpose, we run SPDY with one-shot magnitude pruning without any reconstruction to find profiles (the reported accuracy of course still stitches the resulting profile from the reconstruction database for comparability). As Table \ref{tab:metrics-ablation} shows, this dramatically worsens results, especially at high sparsities where simple one-shot pruning typically completely crashes the model, making it essentially impossible for SPDY to find useful sensitivity coefficients.

In summary, the experiments covered in this section strongly suggest that both the search procedure as well as the reconstruction database are indeed key components of our system and are not easily replaced by simpler means.

\section{Search Procedure Discussion}
\label{app:search}

In the main text, we briefly mentioned how the integration of the DP algorithm makes the search significantly easier. We now provide additional discussion and more details on our experimental setup.

While our search only ever considers solutions that lie directly on the constraint boundary, other approaches \cite{cai2019once, yang2021netadaptv2, li2021brecq} typically have to evaluate many candidates for which the constraint is either violated or not tight (in which case the solution cannot be optimal) in order to make progress. Additionally, the DP solving always ensures proper utilization of the layer-wise speedups. For illustration, a layer with poor acceleration will not be pruned much for most sensitivity values. Similarly, due to the quadratic nature of our errors, most layers will only be pruned to very high sparsities, where speedup curves typically flatten, if this is really necessary to reach the overall speedup target. As a consequence, non-key layers will often have large ranges of acceptable sensitivity scores that all lead to a very similar sparsity choice. This makes finding a good profile in terms of $\mathbf{c}$ easier than by direct optimization in the huge space of all $|S|^L$ possible layer-wise sparsity configurations.\footnote{For example, the architecture search space in \cite{cai2019once} is $\approx 10^{19}$ whereas ours is $\approx 10^{80}$ for the ResNet50 model.}

\begin{figure}[h!]
    \centering
    \includegraphics[width=.9\linewidth]{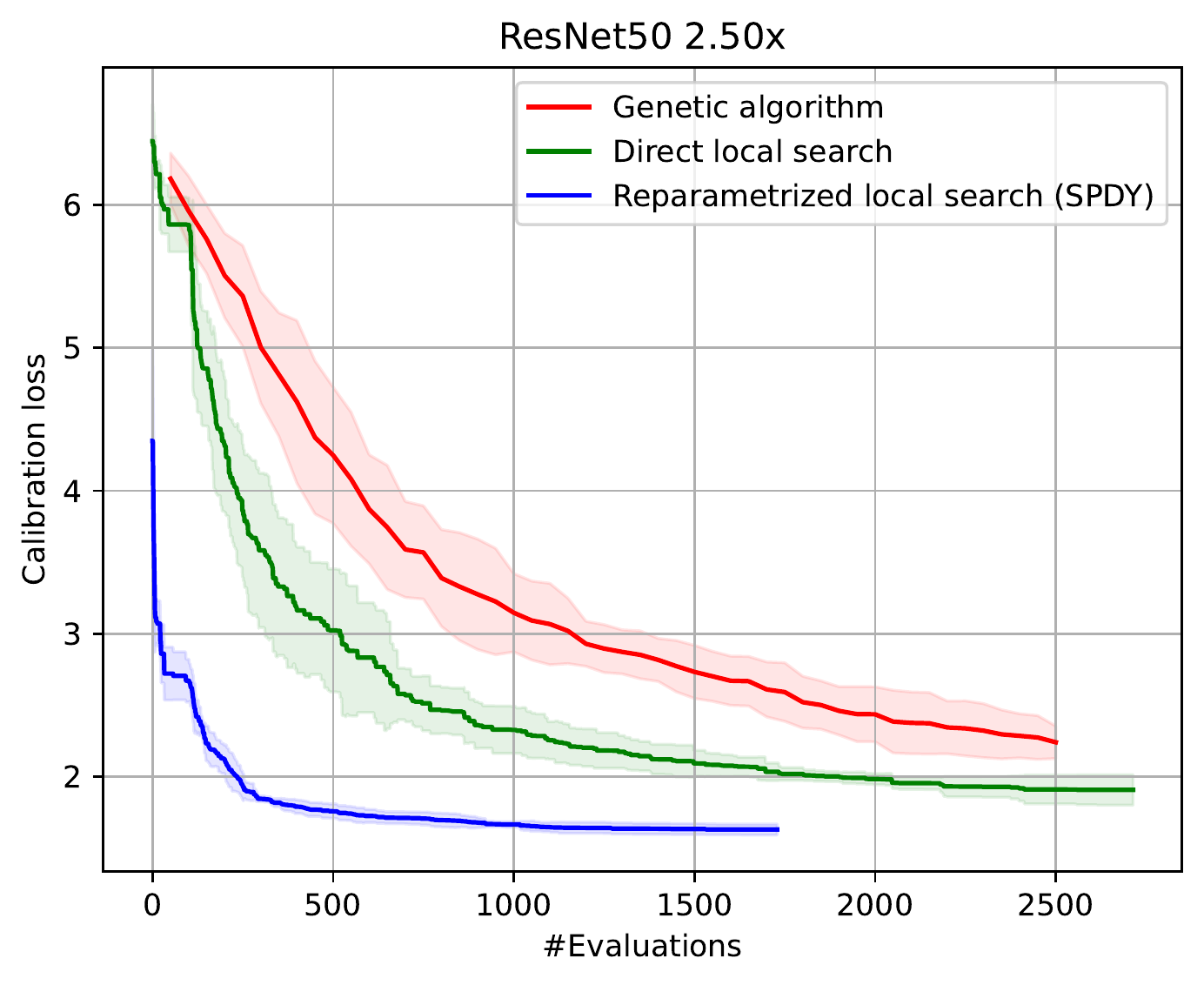}
    \vspace{-10pt}
    \caption{Comparison of our reparametrized unconstrained search and a constrained direct search.}
    \label{fig:search-comparison}
\end{figure}

We now demonstrate these claims empirically by comparing the average over 5 runs of our reparametrized unconstrained search using the DP algorithm, and a direct (constrained) search in the original sparsity space. For comparability, we use exactly the same local search method detailed in Section \ref{sec:error-metric-learning} for the direct search, just with resampling (uniformly in $S$) changes to the current solution until one is found that satisfies the speedup constraint. Additionally, we also test a genetic algorithm, implented exactly as described in \cite{guo2020single} but using $2.5 \times$ more iterations for a fair comparison. Figure \ref{fig:search-comparison} shows the evolution of the objective value relative to the dominating runtime factor, the number of stitched model evaluations (calibration loss computations), for a $2.5\times$ ResNet50 profile.

\begin{figure*}[h]
    \centering
    \begin{subfigure}{.4\linewidth}
      \centering
      \includegraphics[width=\linewidth]{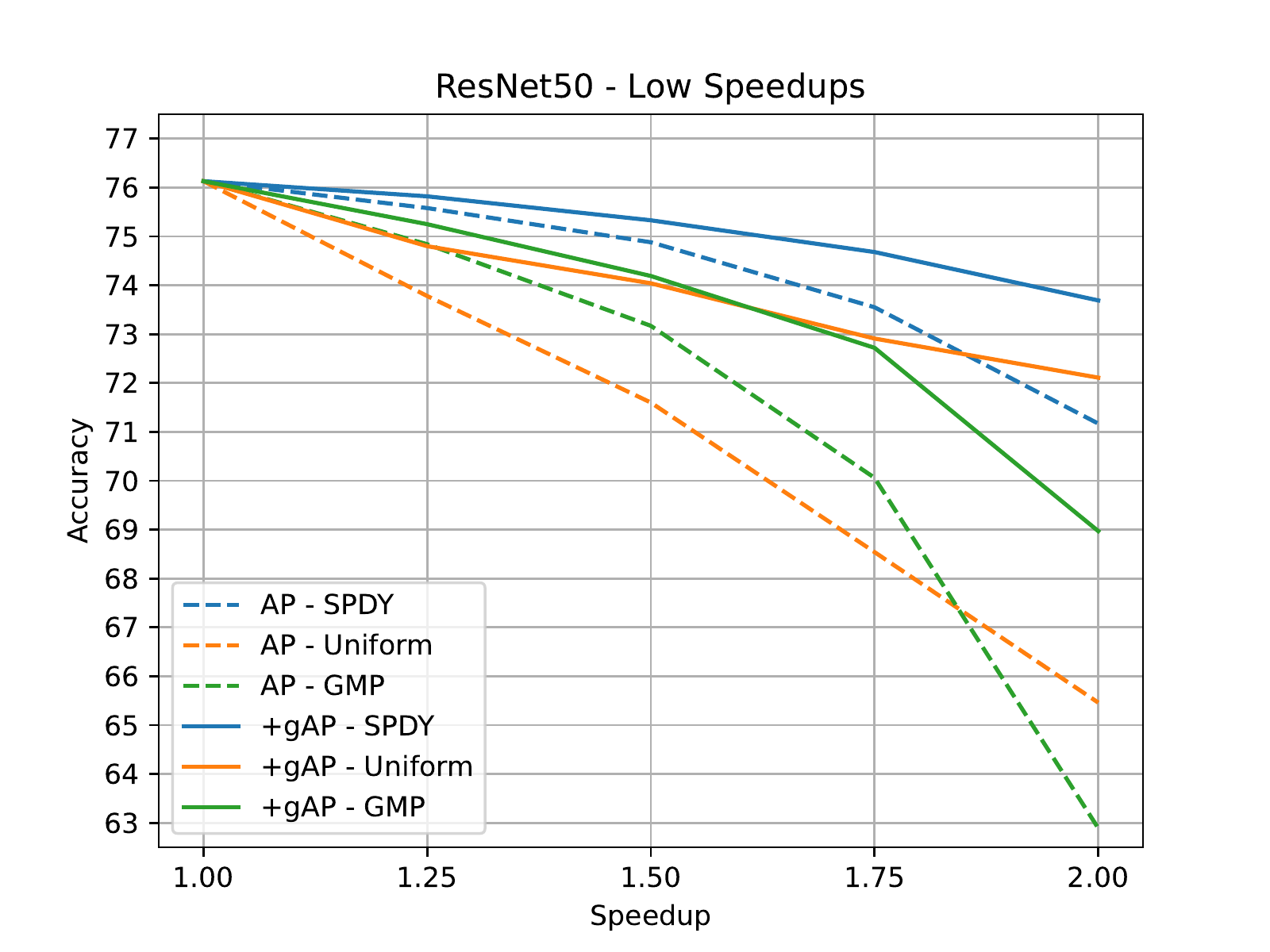}
    \end{subfigure} %
    \begin{subfigure}{.4\linewidth}
      \centering
      \includegraphics[width=\linewidth]{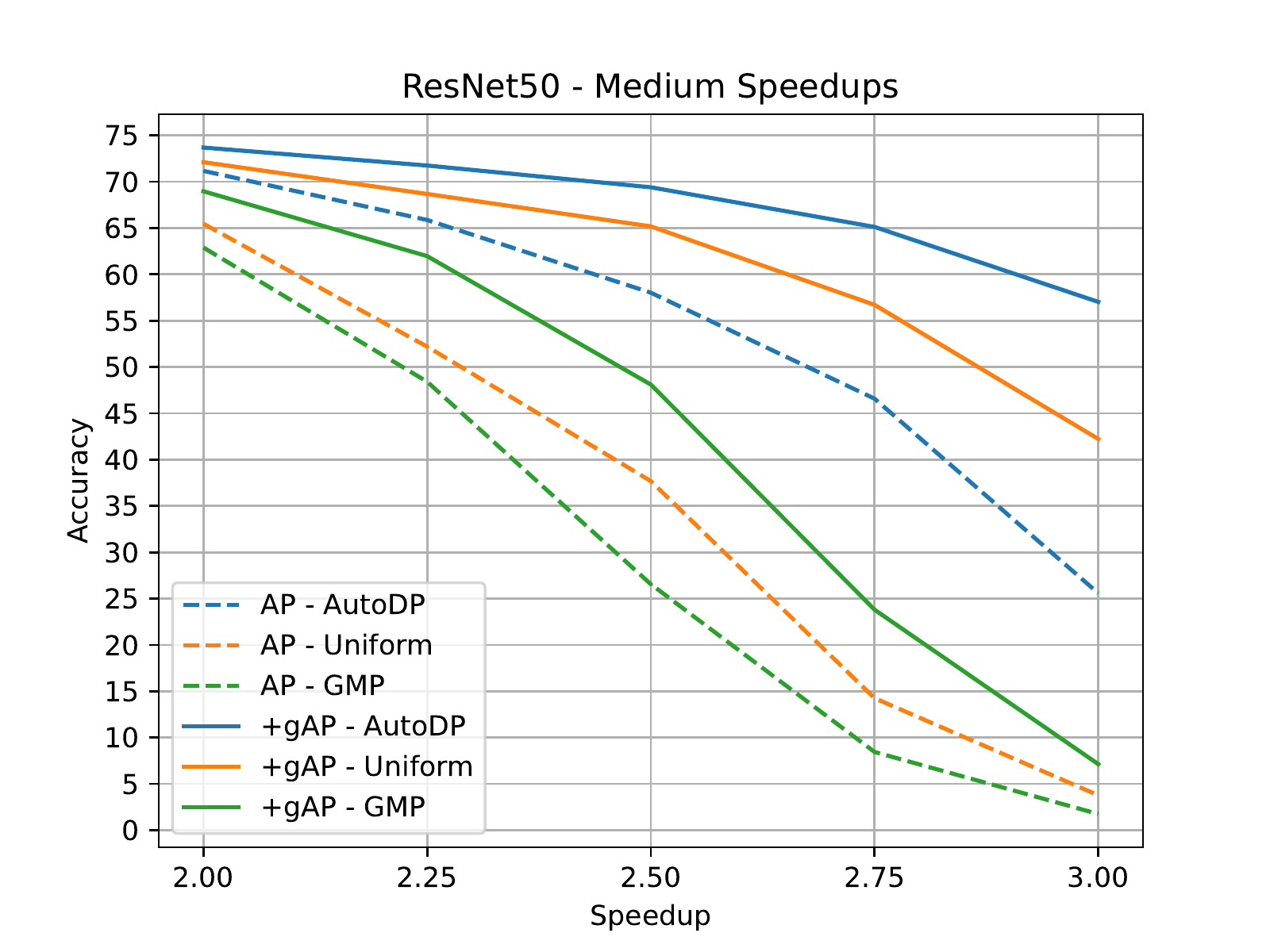}
    \end{subfigure}
    \vspace{-5pt}
    \caption{Accuracy after one-shot pruning vs. speedup for all profile generation methods on ResNet50. For improved visibility, as the y-axes are very different, the plot is split in two showing $1\times$ to $2\times$ speedup on the left and $2\times$ to $3\times$ on the right.}
    \label{fig:rn50-speedup-oneshot}
\end{figure*}

The plot clearly shows how the reparametrized search finds profiles with lower calibration loss, does so with much fewer evaluations ($\approx 10\times$ here) and exhibits significantly reduced variance in solution quality over multiple runs. Overall, this much increased efficiency suggests that the reparametrized DP search could probably also be used in conjunction with better but more expensive evaluation strategies. Interestingly, the genetic algorithm seems to perform noticeably worse than the simple local search. We suspect this is due to fact that in our huge search space, getting stuck in local minima seems to be less of a problem than actually reaching such a minimum quickly. Thus, focusing on mutation evaluations of just the current best solution leads to better results faster than distributing them across several candidates, as is done by genetic programming.

\section{Additional Experiments}
\label{app:additional-experiments}

To investigate the one-shot performance of the different profile generation methods more closely, we prune RN50 to various speedup targets in steps of $0.25$ and plot the results in Figure \ref{fig:rn50-speedup-oneshot}.

First, this plot shows that the graphs of uniform and GMP cross somewhere between $1.5\times$ and $1.75\times$ speedup. We can also observe how the relative order of the accuracies directly after AP stitching is preserved even through big accuracy increases by running global AP. Finally, we can clearly see how the gap between SPDY (after AP but also after gAP) and other methods rapidly grows as the speedup target is increased, up to over $10\%$ accuracy at $3\times$ speedup. Overall, it can be said that SPDY appears to work well in the challenging one-shot setting.

\subsection{Global AdaPrune 2:4}

Additionally, we now present results for post training 2:4 pruning various YOLOv5 models and BERT finetuned on SQuAD in Table \ref{tab:ptq-yolo}. Although the mAP drops for YOLO are quite a bit bigger than the accuracy drops for ResNet, the results are still several points above the next smaller and roughly $2\times$ faster model (a similar acceleration is promised by 2:4 sparsity) and could thus be relevant in practice.

\begin{table}[h]
    \begin{minipage}[c]{\linewidth}
        \centering
        \scalebox{.75}{
            \begin{tabular}{|l|c|c|c|}
                \toprule
                \multirow{2}{*}{Model} & \multirow{2}{*}{Dense} & \multicolumn{2}{c|}{2:4} \\
                & & AP & \textbf{+gAP} \\
                \midrule
                YOLOv5n & 46.20 & 13.40 & \textbf{37.10} \\
                YOLOv5s & 56.40 & 40.70 & \textbf{51.60} \\
                YOLOv5m & 64.20 & 54.00 & \textbf{61.20} \\
                YOLOv5l & 67.40 & 59.80 & \textbf{65.40} \\
                \midrule
                BERT SQuAD & 88.54 & 84.75 & 87.41 \\
                \bottomrule
            \end{tabular}
        }
        \vspace{-5pt}
        \captionof{table}{Global AdaPrune performance for 2:4 pruning the YOLOv5 model family as well as BERT-base on SQuAD. All input and output layers are skipped.}
        \label{tab:ptq-yolo}
    \end{minipage}
\end{table}

\section{Real Timings}
\label{app:real-timings}

\begin{table}[h]
    \begin{minipage}[c]{\linewidth}
        \centering
        \scalebox{.75}{
            \begin{tabular}{|l|c|c|c|}
                \toprule
                Model & Target & CPU & Real \\
                \midrule
                ResNet50 & $1.00\times$ & AMD & 0.478s -- $1.00\times$ \\
                ResNet50 & $2.00\times$ & AMD & 0.234s -- $2.04\times$ \\
                ResNet50 & $2.50\times$ & AMD & 0.178s -- $2.69\times$ \\
                ResNet50 & $3.00\times$ & AMD & 0.143s -- $3.34\times$ \\
                ResNet50 & $3.50\times$ & AMD & 0.121s -- $3.95\times$ \\
                \midrule
                MobileNetV1 & $1.00\times$ & Intel & 0.045s -- $1.00\times$ \\
                MobileNetV1 & $1.50\times$ & Intel & 0.031s -- $1.45\times$ \\
                \midrule
                YOLOv5s & $1.00\times$ & Intel & 0.641s -- $1.00\times$ \\
                YOLOv5s & $1.50\times$ & Intel & 0.449s -- $1.43\times$ \\
                YOLOv5s & $1.75\times$ & Intel & 0.380s -- $1.68\times$ \\
                YOLOv5m & $1.00\times$ & Intel & 1.459s -- $1.00\times$ \\
                YOLOv5m & $1.75\times$ & Intel & 0.848s -- $1.72\times$ \\
                YOLOv5m & $2.00\times$ & Intel & 0.725s -- $2.01\times$ \\
                \midrule
                BERT SQuAD & $1.00\times$ & Intel & 0.969s -- $1.00\times$ \\
                BERT SQuAD & $3.00\times$ & Intel & 0.320s -- $3.03\times$ \\
                BERT SQuAD & $3.50\times$ & Intel & 0.271s -- $3.58\times$ \\
                BERT SQuAD & $4.00\times$ & Intel & 0.223s -- $4.35\times$ \\
                \bottomrule
            \end{tabular}
        }
        \vspace{-5pt}
        \captionof{table}{Real timings of the SPDY profiles in Table \ref{tab:gradual-results} using the DeepSparse v0.9.1 engine. All timings are for batchsize 64, except for BERT which uses batchsize 16.}
        \label{tab:real-timings}
    \end{minipage}
\end{table}

Table \ref{tab:real-timings} shows the real timings of the final pruned models resulting from the SPDY profiles in our gradual pruning experiments in Section \ref{sec:gradual-pruning}. We can see that the speedup predictions of the additive model are in most cases very accurate, especially considering typical performance fluctuations with varying operating temperature and that even different CPUs of the same type can have small performance differences. Only for the higher speedup ResNet50 models and $4.00\times$ BERT, the true speedups are noticeably underestimated. This is most likely due to special engine optimizations (we run with all of them turned on here) that are not active in the layer-wise timing mode. With sufficient knowledge about the engine internals, it would probably be possible to account for such effects in the DP algorithm. However, optimizing our methods for one particular inference engine was not the goal of this work.

\section{Overall Sparsities}
\label{app:sparsity}

As our focus are real model speedups, our results in the main paper are all given with respect to those rather than parameter counts. Nevertheless, for completeness we now provide overall sparsity values (with respect to all pruned layers) corresponding to our main gradual pruning results in Table \ref{tab:gradual-results}.

\begin{table}[h]
    \begin{minipage}[c]{\linewidth}
        \centering
        \scalebox{.75}{
            \begin{tabular}{|l|c|c|c|c|c|}
                \toprule
                Model & Speed. & CPU & SPDY & Uni. & GMP \\
                \midrule
                ResNet50 & $2.00\times$ & AMD & 70.31 & 80.54 & 85.77 \\
                ResNet50 & $2.50\times$ & AMD & 86.15 & 88.33 & 92.30 \\
                ResNet50 & $3.00\times$ & AMD & 91.78 & 93.01 & 96.24 \\
                ResNet50 & $3.50\times$ & AMD & 95.22 & 96.58 & 98.26 \\
                \midrule
                MobileNetV1 & $1.50\times$ & Intel & 49.13 & 65.88 & 77.16 \\
                \midrule
                YOLOv5s & $1.50\times$ & Intel & 59.83 & 70.69 & 70.84 \\
                YOLOv5s & $1.75\times$ & Intel & 85.16 & 85.68 & 91.60 \\
                YOLOv5m & $1.75\times$ & Intel & 79.81 & 82.42 & 88.20 \\
                YOLOv5m & $2.00\times$ & Intel & 90.42 & 91.42 & 94.29 \\
                \midrule
                BERT SQuAD & $3.00\times$ & Intel & 82.24 & 84.14 & 83.98 \\
                BERT SQuAD & $3.50\times$ & Intel & 88.89 & 90.49 & 90.18 \\
                BERT SQuAD & $4.00\times$ & Intel & 94.15 & 94.86 & 95.29 \\
                \bottomrule
            \end{tabular}
        }
        \vspace{-5pt}
        \captionof{table}{Overall sparsities of profiles corresponding to Table \ref{tab:gradual-results}, calculated with respect to all pruned layers.}
        \label{tab:overall-sparsities}
    \end{minipage}
\end{table}

As can be seen in Table \ref{tab:overall-sparsities}, SPDY profiles generally achieve the same speedup with a lower overall sparsity. Similarly, the relative difference in the number of remaining weights tends to increase for higher speedups. This is expected as our method directly takes execution information into account and can thus e.g. prioritize pruning layers that provide good speedups while leaving more weights on other (perhaps bigger) layers with worse acceleration behavior. However, we note that our techniques are quite general and can be easily adapted (by replacing the timings in the DP problem with the number of remaining parameters) to optimize for overall sparsity.

\section{Profile Visualizations}
\label{app:profiles}

\begin{figure*}[h]
    \centering
    \begin{subfigure}{.45\linewidth}
        \centering
        \includegraphics[width=\linewidth]{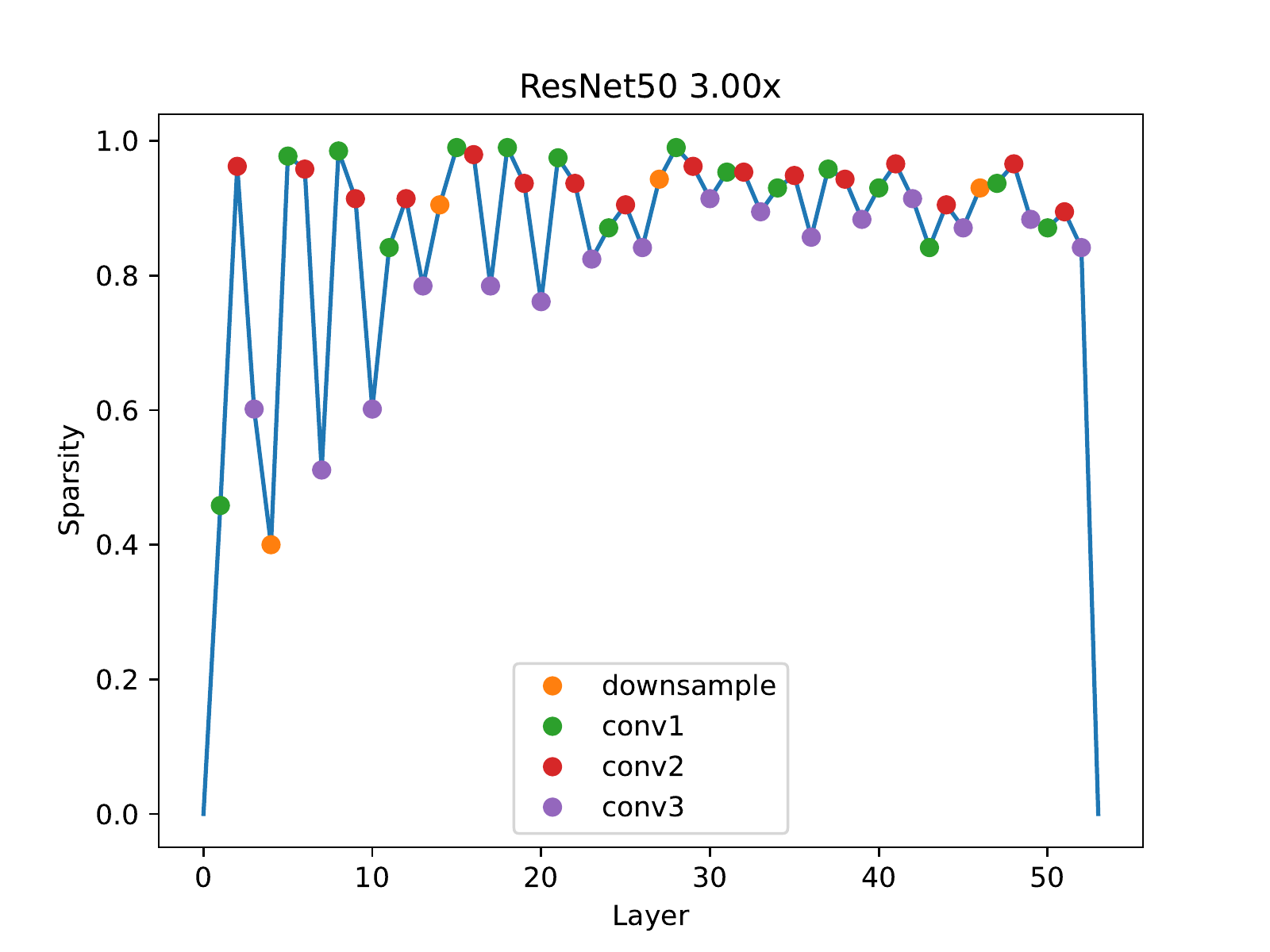}
    \end{subfigure} %
    \begin{subfigure}{.45\linewidth}
        \centering
        \includegraphics[width=\linewidth]{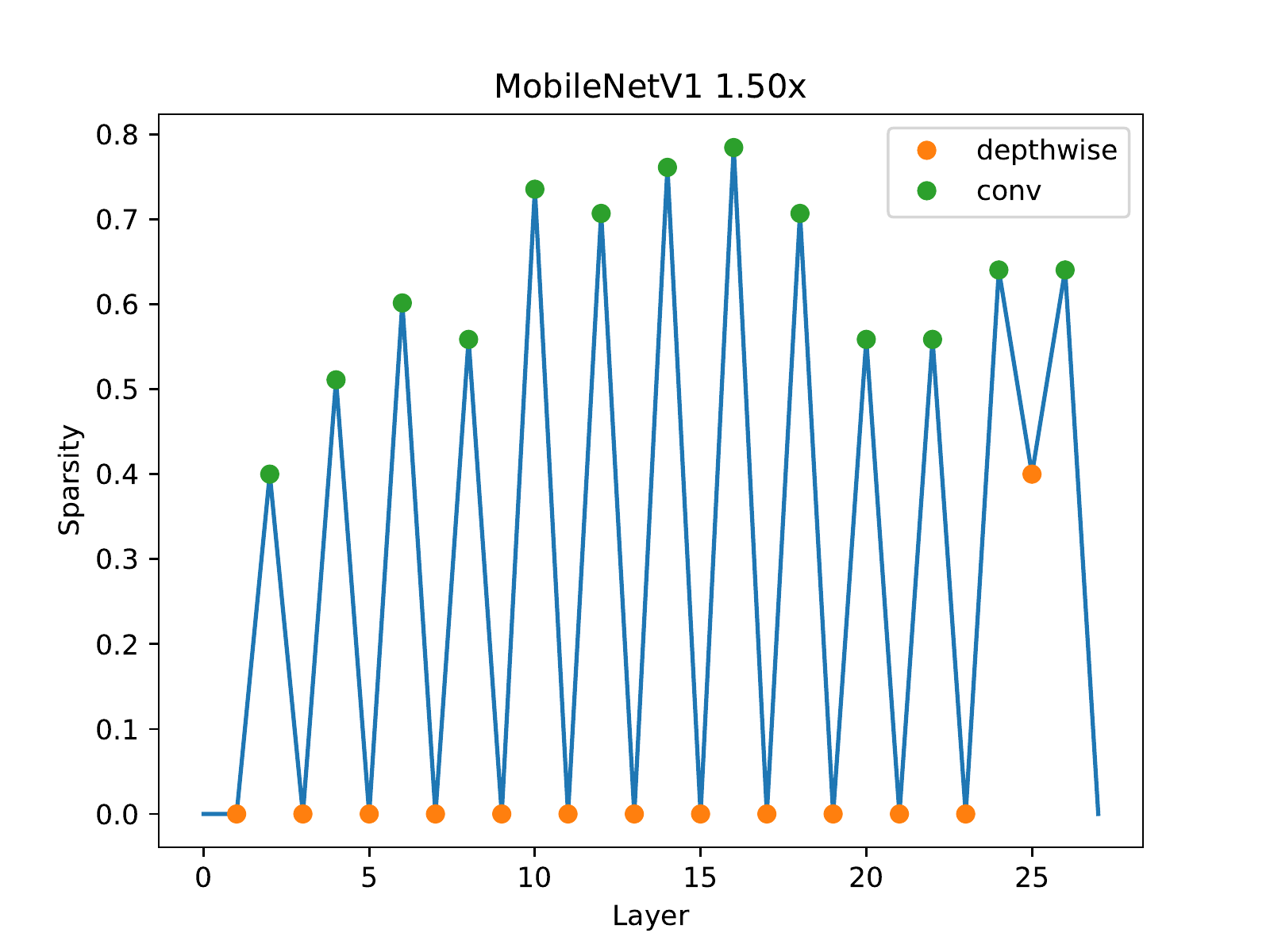}
    \end{subfigure} %
    \vspace{-10pt}
    \begin{subfigure}{.45\linewidth}
        \centering
        \includegraphics[width=\linewidth]{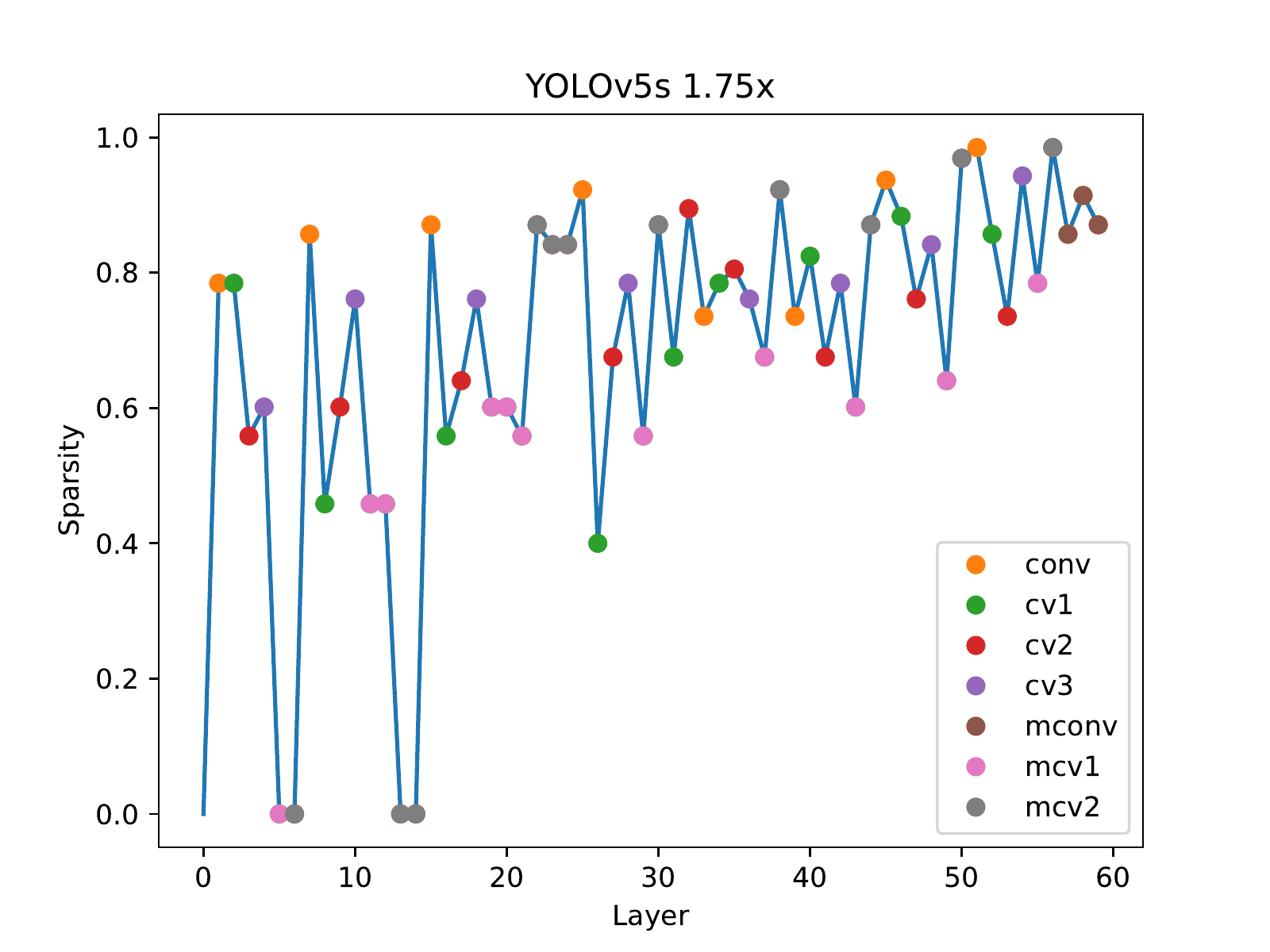}
    \end{subfigure} %
    \begin{subfigure}{.45\linewidth}
        \centering
        \includegraphics[width=\linewidth]{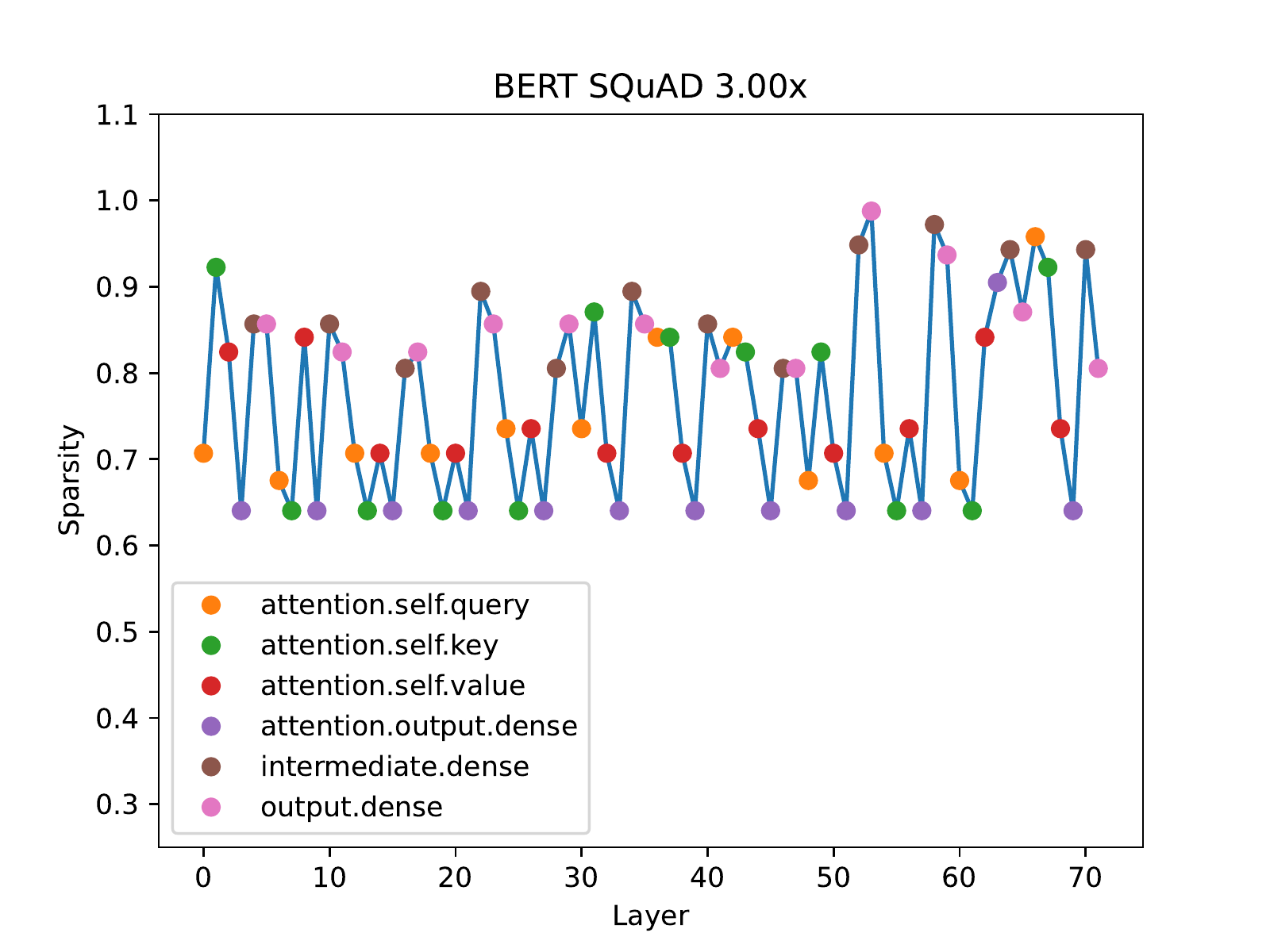}
    \end{subfigure}
    \caption{Sample visualizations of SPDY profiles.}
    \label{fig:profile-visualizations}
\end{figure*}

Figure \ref{fig:profile-visualizations} displays visualizations of some of the SPDY profiles in Section \ref{sec:gradual-pruning}. We will now analyze those briefly.
Starting with ResNet50, we can see how the last convolution in a residual-block is typically pruned less than the others, with this effect being most pronounced in the early layers. Further, we can observe how the first conv is pruned more than the second one early on with the roles seemingly switching in the later layers.
Next, for MobileNetV1, we can see that SPDY keeps all but the very last depth-wise convolution dense since those allow almost no acceleration while at the same being very sensitive. For the standard convolutions, SPDY seems to do the most pruning in the middle layers. YOLOv5s is a quite complex model and features also a correspondingly complex profile. We can see that the first conv of the pathway leading to the eventual output maps ``mconv1'' is typically pruned less than the other layers while the convolution following it ``mconv2'' is typically amongst the most strongly pruned ones. Additionally, convolutions not in a residual block ``conv'' are also pruned quite heavily in most cases. At last, one can notice that the 3 output layers are pruned to relatively high levels, verifying that those should not be skipped like for other models. In the BERT profile, the first projection after attention ``attention.output.dense'' is generally not pruned very strongly and the query and value matrices typically hover around the middle in terms of the assigned sparsities. Meanwhile, the fully-connected layers ``intermediate.dense'' and ``output.dense'' are usually amongst the most strongly pruned parts, which is to be expected as those make up a big portion of the overall run-time. All in all, the profiles found by SPDY exhibit various interesting patterns and carefully balance the layer-wise speed-sensitivity trade-offs in ways that seem very difficult to derive manually.

\end{document}